\documentclass[10pt,journal,cspaper,compsoc]{IEEEtran}

\ifCLASSOPTIONcompsoc
  \usepackage[nocompress]{cite}
\else
  \usepackage{cite}
\fi

\ifCLASSINFOpdf
   \usepackage[pdftex]{graphicx}
   \graphicspath{{../pdf/}{../jpeg/}}
   \DeclareGraphicsExtensions{.pdf,.jpeg,.png,.jpg}
\else
   \usepackage[dvips]{graphicx}
   \graphicspath{{../eps/}}
   \DeclareGraphicsExtensions{.eps}
\fi

\usepackage{hyperref}
\usepackage{amsmath}
\usepackage{algorithm}
\usepackage{algorithmic}
\usepackage{amsfonts}
\usepackage{wrapfig}

\ifCLASSOPTIONcompsoc
  \usepackage[caption=false,font=footnotesize,labelfont=sf,textfont=sf]{subfig}
\else
  \usepackage[caption=false,font=footnotesize]{subfig}
\fi

\usepackage{stfloats}

\usepackage{amssymb}

\usepackage{cleveref}
\usepackage{xcolor}
\usepackage{adjustbox}
\usepackage{xspace}
\usepackage{subcaption}
\usepackage{multirow}
\usepackage{xcolor}
\definecolor{myorange}{RGB}{236,166,128} % Define a custom orange color
\definecolor{myblue}{RGB}{39,131,177} % Define a custom orange color
\usepackage{tikz} % For drawing lines
\newcommand{\name}{\textit{WonderHuman}\xspace}

\usepackage{ragged2e}

\begin{document}

\title{
WonderHuman: Hallucinating Unseen Parts in Dynamic 3D Human Reconstruction
}

\author{Zilong~Wang,~Zhiyang~Dou,~Yuan~Liu,~Cheng~Lin,~Xiao~Dong,~Yunhui~Guo,~Chenxu~Zhang,\\~Xin~Li,~Wenping~Wang,~Xiaohu~Guo% <-this % stops a space
\IEEEcompsocitemizethanks{\IEEEcompsocthanksitem Z. Wang, Y. Guo, C. Zhang and X. Guo are with the Department of Computer Science, The University of Texas at Dallas, Richardson, Texas.
\IEEEcompsocthanksitem Z. Dou and C. Lin are with the Computer Graphics Group, The University of Hong Kong, Pokfulam, Hong Kong.
\IEEEcompsocthanksitem Y. Liu is with the School of Engineering, The Hong Kong University of Science and Technology, Clear Water Bay, Hong Kong.
\IEEEcompsocthanksitem X. Dong is with Guangdong Provincial/Zhuhai Key Laboratory of IRADS, Beijing Normal-Hong Kong Baptist University, Zhuhai, China.
\IEEEcompsocthanksitem X. Li and W. Wang are with the Department of Computer Science \& Engineering, Texas A\&M University, College Station, Texas.

\IEEEcompsocthanksitem Corresponding Author: X. Guo, Email: xguo@utdallas.edu; Z. Dou, Email: frankzydou@gmail.com
}
}

\markboth{Submission to IEEE Transactions on Visualization and Computer Graphics}%
{Z. Wang \MakeLowercase{\textit{et al.}}:}

\IEEEtitleabstractindextext{%
\begin{abstract}
\justifying In this paper, we present \textit{WonderHuman} to reconstruct dynamic human avatars from a monocular video for high-fidelity novel view synthesis. 
Previous dynamic human avatar reconstruction methods typically require the input video to have full coverage of the observed human body. However, in daily practice, one typically has access to limited viewpoints, such as monocular front-view videos, making it a cumbersome task for previous methods to reconstruct the unseen parts of the human avatar. To tackle the issue, we present \textit{WonderHuman}, which leverages 2D generative diffusion model priors to achieve high-quality, photorealistic reconstructions of dynamic human avatars from monocular videos, including accurate rendering of unseen body parts. Our approach introduces a Dual-Space Optimization technique, applying Score Distillation Sampling (SDS) in both canonical and observation spaces to ensure visual consistency and enhance realism in dynamic human reconstruction. Additionally, we present a View Selection strategy and Pose Feature Injection to enforce the consistency between SDS predictions and observed data, ensuring pose-dependent effects and higher fidelity in the reconstructed avatar. In the experiments, our method achieves SOTA performance in producing photorealistic renderings from the given monocular video, particularly for those challenging unseen parts. The project page and source code can be found at \url{https://wyiguanw.github.io/WonderHuman/}.%The code will be made publicly accessible upon publication.

\end{abstract}

}

\maketitle

\IEEEdisplaynontitleabstractindextext

\IEEEpeerreviewmaketitle

\IEEEraisesectionheading{\section{Introduction}\label{sec:introduction}}
\IEEEPARstart{V}{irtual} avatars have been a key focus in computer vision, graphics, and VR/AR technologies due to their wide applications such as gaming, entertainment, communication, and telepresence. However, reconstructing high-fidelity avatars that faithfully represent human appearance, shape, and dynamics remains a formidable challenge, particularly when confronted with ubiquitous monocular video with highly limited viewpoints.

Existing avatar reconstruction methods have difficulty in reconstructing unseen parts of the human body.
Previous methods~\cite{peng2021neural,peng2024animatable,su2022danbo,wang2022arah} typically rely on dense, synchronized multi-view inputs for the avatar reconstruction task. Recent advancements in implicit neural radiance fields~\cite{mildenhall2021nerf,peng2021neural,su2022danbo,wang2022arah} and 3D Gaussian Splatting~\cite{kerbl3Dgaussians,lei2023gart,yin2023humanrecon,zheng2024gpsgaussian} have explored the high-fidelity reconstruction of both geometry and appearance of dynamic human bodies from relatively sparse multi-view videos. To reconstruct from monocular videos, other recent methods~\cite{weng2022humannerf,su2021nerf,su2023npc,yu2023monohuman,huang2023efficient,instant_nvr,hu2023gaussianavatar} reconstruct dynamic avatars by animating them within a canonical space derived from observation spaces using video frames. These works enable learning the inter-frame deformation to reconstruct a completed human avatar from the monocular videos. However, these methods still require the video to have full-view coverage of the human body, and typically fail to reconstruct unseen parts in the monocular video. Unfortunately, one often only has access to partial-view videos with limited viewpoints, such as front-view videos, leaving most parts of the human body unseen. Reconstructing these occluded parts thus poses a significant challenge for current methodologies.

To address this challenge, we introduce \name to achieve high-quality avatar reconstruction from partial-view monocular videos. 

The key idea of \name is to hallucinate the unseen parts of the human using the generative prior encoded by large-scale image diffusion models such as Zero123~\cite{liu2023zero1to3}. The hallucinations are then combined with a Gaussian Splatting~\cite{kerbl3Dgaussians}-based dynamic human reconstruction framework to obtain a full-body avatar. 

However, combining diffusion-based generative priors in dynamic human reconstruction is not a trivial task with two outstanding challenges. First, the existing image diffusion generative models are designed mainly to produce single-view \textit{ static} images. Thus, maintaining visually accurate generated content and consistency across frames for \textit{dynamic} human bodies using these generative priors is challenging. For instance, unrealistic artifacts such as blurs often appear when animating the generated bodies. Some existing works~\cite{ho2024sith,huang2023tech,albahar2023single} can produce human bodies from single-view images using diffusion models, but they fail to handle dynamic cases (See Appendix B.1 for more details). Second, it is challenging to ensure that the occluded or invisible portions of the human body generated by diffusion models are consistent with the observed visible parts. Any inconsistency between these generated and visible segments can significantly deteriorate the rendering quality of the human avatar, leading to visually incoherent results. 

\IEEEpubidadjcol
To tackle these issues, we present \name for high-quality dynamic human reconstruction from monocular videos. We propose leveraging the generative priors embedded in a 2D diffusion model, trained on condensed images, to infer the unseen parts of the 3D human through distillation during reconstruction. We further introduce a novel Dual-space Optimization method to ensure visual plausibility and consistency for dynamic human representations. Our Dual-space Optimization utilizes Score Distillation Sampling (SDS)~\cite{poole2022dreamfusion} in both the canonical and observation spaces. This approach is designed to bridge the gap between observation and canonical spaces by effectively fusing detailed appearance information captured from multi-pose observations into a consistent canonical representation. This process not only preserves spatial fidelity but also enhances temporal coherence (frame-by-frame) by modeling dynamic changes over time, leading to more stable and accurate reconstructions of animatable human avatars.

Moreover, a view selection strategy and a pose feature injection approach are employed to reconcile conflicts between the SDS predictions and the given information and fuse pose-dependent effects, enhancing dynamic synthesis and overall avatar fidelity.

We conduct extensive experiments to validate the effectiveness of our method across broad benchmarks including ZJU-Mocap dataset~\cite{peng2021neural}, Monocap dataset~\cite{peng2024animatable}, MVHumanNet~\cite{xiong2024mvhumannet}, and In-the-wild dataset~\cite{zhou2019dance}. 
Compared to state-of-the-art methods~\cite{weng2022humannerf,instant_nvr,shao2024splattingavatar,moon2024exavatar,lee2024gtu}, \name produces higher-quality photorealistic renderings of reconstructed human avatars, particularly in rendering visually plausible content for previously unseen parts of the human body. To summarize, our contributions are as follows:

\begin{itemize}

    \item 
    We propose a novel framework named \name that leverages 2D generative diffusion priors to achieve high-quality, photorealistic reconstruction of dynamic humans from monocular videos, including accurate rendering of unseen body parts. 
    \item 
    We introduce Dual-space Optimization to ensure visual consistency and enhance realism throughout the dynamic reconstruction process.
    \item 
   We present a view selection strategy alongside pose feature injection to resolve conflicts between SDS predictions and observed data, ensuring pose-dependent effects and higher fidelity in the reconstructed avatar.

\end{itemize}

% \vspace{-5mm}
\section{Related work}

\subsection{Video-based Human Avatar Reconstruction}
Recently, video-based avatar reconstruction methods primarily rely on regression-based approaches~\cite{he2020geo,he2021arch++,huang2020arch,saito2019pifu,saito2020pifuhd,xiu2022icon,xiu2023econ,zheng2021pamir} or the explicit tracking of human bodies~\cite{habermann2020deepcap,habermann2019livecap,xu2018monoperfcap,alldieck2018video,casado2022pergamo,guo2021human,moon20223d}. Since the prosperity of Neural Rendering~\cite{mildenhall2021nerf}, many works~\cite{li2022tava,li2023posevocab,liu2021neural,peng2021neural,peng2024animatable,wang2022arah,zheng2022structured,jiang2022selfrecon,jiang2022neuman,su2021nerf,su2022danbo,su2023npc,weng2022humannerf,guo2023vid2avatar,yu2023monohuman} try to combine neural representations with human reconstructions. These methods associate implicit neural fields with human templates such as Skinned Multi-Person Linear model (SMPL)~\cite{SMPL:2015}, a widely used parametric 3D human body model that represents pose and shape using a learned mesh deformation framework. While neural representations have strong representation ability, they are slow in training. But other works~\cite{zhao2022high,huang2023efficient,zheng2023pointavatar,instant_nvr} additionally introduced explicit representations like meshes~\cite{zhao2022high,huang2023efficient}, and points~\cite{zheng2023pointavatar} to improve its efficiency. Yet, achieving high-quality reconstruction results using neural radiance fields still requires neural networks that are expensive to train and render. Recently, Gaussian Splatting~\cite{kerbl3Dgaussians} has emerged as a prominent technology, as it efficiently represents and renders complex scenes with reduced training time, without compromising quality for speed. Many recent works~\cite{lei2023gart,li2023animatable,yin2023humanrecon,hu2023gauhuman,jiang2023hifi4g,li2023human101,hu2023gaussianavatar,qian20233dgsavatar,zheng2024gpsgaussian,liu24-GVA,reloo,kim2024moconerf,pramishp2024iHuman} try to combine Gaussian Splatting in the avatar reconstruction, which allows efficient avatar rendering in real-time. In this paper, we focus on reconstructing human avatars from monocular video. In GaussianAvatar~\cite{hu2023gaussianavatar}, 3D Gaussians are integrated with SMPL~\cite{SMPL:2015} to explicitly represent humans in various poses and clothing styles. SplattingAvatar~\cite{shao2024splattingavatar} embeds Gaussians onto human triangle meshes, forming a hybrid representation that significantly enhances rendering speed. Furthermore, ExAvatar~\cite{moon2024exavatar} extends this representation to reconstruct animatable hand poses and facial expressions. Building on prior advances, GPS-Gaussians+~\cite{GPS-Gaussian+} extend the capability of GPS-Gaussian~\cite{zheng2024gpsgaussian} from dense-view input to sparse-view input. TaoAvatar~\cite{Chen_2025_CVPR} delivers real-time, high-fidelity full-body avatars capable of running on mobile and wearable devices, bridging the gap between quality and efficiency. And MIMO~\cite{Men_2025_CVPR} achieves avatar generation, motion generation, and user interaction in a unified framework. Vid2Avatar~\cite{Guo_2025_CVPR} addresses overfitting issues by using a universal model for avatar reconstruction from monocular video. However, those methods require the input video to have full-view coverage of the human body and fail to generate unseen parts in the monocular video.

\subsection{Diffusion Models for Human Avatars}

Pioneer works in avatar generation~\cite{hong2022avatarclip} resort to generate avatars from CLIP features~\cite{radford2021learning}. Recently, diffusion models~\cite{ho2020denoising} show strong ability in learning complex data distributions for data generation. Some works~\cite{zhang2023avatarverse,huang2024dreamwaltz,kolotouros2024dreamhuman,zeng2023avatarbooth,cao2023dreamavatar,jiang2023avatarcraft,zhang2023styleavatar3d,kim2023chupa,cao2023guide3d,mendiratta2023avatarstudio,liao2023tada,wang2023disentangled,liu2023humangaussian,xu2023seeavatar,wang2023humancoser} directly extend the SDS loss~\cite{poole2022dreamfusion} to generate human avatars from text prompts. MVHuman~\cite{jiang2023mvhuman} extends this framework to generate human avatars through multiview diffusion, while HumanNorm~\cite{huang2023humannorm} integrates it with normal map generation. Additionally, HumanNorm~\cite{hu2023humanliff} directly enables 3D human generation, benefiting from tri-plane features. Some other works~\cite{huang2023tech,svitov2023dinar,alldieck2022photorealistic,zhang2023humanref,albahar2023single,ho2024sith,chen2024ghgsingleview,pan2024hsgsingleimagehuman,sun2024occfusion} generate a completed human avatar from a single-view image using diffusion models. 
While these single-view avatar generation techniques produce avatars from single images, directly extending them to generate dynamic humans from monocular videos results in poor rendering quality for dynamic human actions. In contrast, our approach leverages the SDS loss to inpaint the unseen parts of the dynamic human body from a monocular video, with careful consideration of time coherence, consistency, and dynamics.

\section{Preliminaries }
\label{sec:Preliminaries}

\subsection{3D Gaussian Splatting}
3D Gaussian splatting~\cite{kerbl3Dgaussians} is an explicit scene representation that allows high-quality real-time rendering. The given scene is represented by a set of static 3D Gaussians, which are parameterized as follows: Gaussian center $x\in {\mathbb{R}^3}$, color $c\in {\mathbb{R}^3}$, opacity $\alpha\in {\mathbb{R}}$, spatial rotation in the form of quaternion $q\in {\mathbb{R}^4}$, and scaling factor $s\in {\mathbb{R}^3}$. Given these properties, the rendering process is represented as:

\begin{equation}
  I = Splatting(x, c, s, \alpha, q, r),
  \label{eq:splattingGA}
\end{equation}
where $I$ is the rendered image, $r$ is a set of query rays crossing the scene, and $Splatting(\cdot)$ is a differentiable rendering process. We refer readers to Kerbl et al.'s paper~\cite{kerbl3Dgaussians} for the details of Gaussian splatting.

\subsection{Score Distillation Sampling}
Score Distillation Sampling (SDS)~\cite{poole2022dreamfusion} builds a bridge between diffusion models and 3D representations. In SDS, the noised input is denoised in one time-step, and the difference between added noise and predicted noise is considered SDS loss, expressed as:

\begin{equation}
    \mathcal{L}_{\text{SDS}}(I_{\Phi}) \triangleq \mathbb{E}_{t,\epsilon} \left[ w(t) \left( \epsilon_{\phi}(z_t, y, t) - \epsilon \right) \frac{\partial I_{\Phi}}{\partial \Phi} \right],
  \label{eq:SDSObservGA}
\end{equation}

where the input $I_{\Phi}$ represents a rendered image from a 3D representation, such as 3D Gaussians, with optimizable parameters $\Phi$. $\epsilon_{\phi}$ corresponds to the predicted noise of diffusion networks, which is produced by incorporating the noise image $z_t$ as input and conditioning it with a text or image $y$ at timestep $t$. The noise image $z_t$ is derived by introducing noise $\epsilon$ into $I_{\Phi}$ at timestep $t$. The loss is weighted by the diffusion scheduler $w(t)$.

\section{Method}
\label{sec:method}

\begin{figure*}
\centering
\includegraphics[width=1\textwidth]{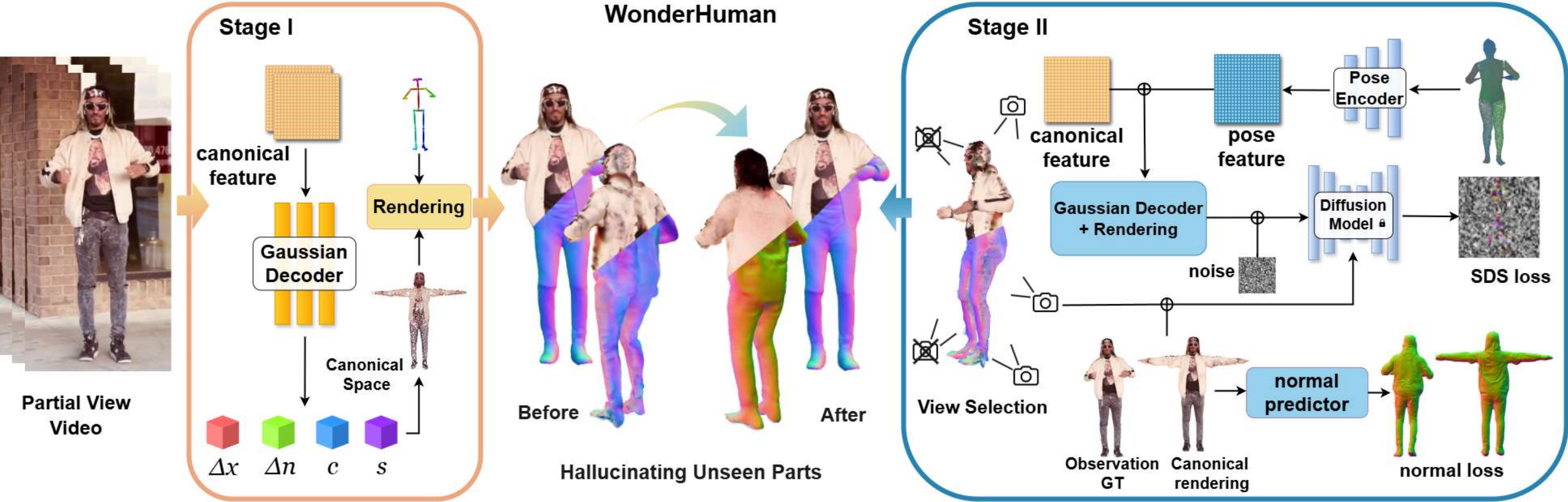} 

\caption{Overview of WonderHuman. (1) In stage I, we reconstruct 3D Gaussians and appearances for visible human parts from partial-view videos. We start with optimizable feature vectors named canonical features capturing human geometry and appearance in a canonical space. Then, we use a Gaussian Decoder to predict Gaussian parameters and combine the Linear Blend Skinning (LBS) function with the Gaussian Splatting to render the dynamic 3D human in the observation space. (2) In Stage II, we hallucinate the invisible parts of the avatar using a Dual-space Optimization technique. We render images of the human avatar from various novel viewpoints and apply an SDS loss to learn the unseen appearances. Additionally, a normal predictor is utilized to generate normal maps that guide geometry reconstruction, while View Selection and Pose Feature Injection strategies are employed to ensure consistent appearance fusion.}

\label{fig:pipline}

\end{figure*}

Given a monocular video as the input, our goal is to reconstruct a high-quality animatable 3D human avatar including both visible and invisible parts. In \name, we employ a dynamic 3D human Gaussian representation, equipped with a generative diffusion model as hallucination prior, which produces a controllable 3D human avatar viewable from any angle. An overview of our method can be found in Fig.~\ref{fig:pipline}.

\subsection{Stage I: Visible Appearance Reconstruction}
\subsubsection{Prediction of Gaussian Parameters} 
\label{gaussianDecoder}
In the first stage, we reconstruct the visible geometry and appearance of an animatable human avatar from a partial-view monocular video. To achieve detailed and high-fidelity reconstructions, building on GaussianAvatar~\cite{hu2023gaussianavatar}, we propose integrating normal information into the Gaussian decoder~\cite{hu2023gaussianavatar}. This improved decoder is used to establish a functional mapping from the underlying geometry of the human to various attributes of 3D Gaussians. And those Gaussians are initialized on the surfaces of SMPL~\cite{SMPL:2015} body in canonical space. Then, we have:

\begin{equation}
  (\Delta x,\Delta n, c,s)=G_{\theta}([S,S]),
  \label{eq:gaussiandecoderstage1}
\end{equation}
where $\theta$ represents optimizable parameters for the Gaussian decoder $G_{\theta}$, and \( S \) represents the features in the canonical space. 
% \TODO{give definition and explanation of this S, channel...}
The canonical feature $S$ is an optimizable tensor, randomly initialized and optimized during training to capture texture and geometry features in canonical space. The size of $S$ is $(128 \times 128)$, and it is concatenated with itself as input of Gaussian decoder $G$. This ensures that the input channel of $G$ remains $(2 \times 128 \times 128)$ during pose feature injection in Stage II (Sec.~\ref{poseInject}).
This decoder $G$ predicts 3D center offset $\Delta{x}$, along with color and scale factors, denoted as $c$ and $s$ respectively. Additionally, it predicts normal offset \(\Delta n\) that is applied to the initial SMPL normals, to capture the intrinsic geometric details. We set the opacity $\alpha$ and 3D rotation $q$ are set to fixed values of $1$ and $(1,0,0,0)$ respectively, to make the network focus more on the geometry information. %enhance geometry reconstruction accuracy.

\subsubsection{Dynamic Human Rendering} 
\label{LBS&rendering}
To render the avatar in observation space, we seamlessly combine the Linear Blend Skinning function with the Gaussian Splatting~\cite{kerbl3Dgaussians} process to deform the avatar from canonical space to observation space:
\begin{equation}
  I_{rgb}=Splatting(x_o,c,Q,r),
  \label{eq:splatting}
\end{equation}
\begin{equation}
  x_o = T_{lbs}(x_c,p,w),
  \label{eq:LBS}
\end{equation}
where $I_{rgb}$ represents the final rendered image. The final canonical Gaussian position $x_c$ is the sum of the initial position $x$ and the predicted offset $\Delta x$. The LBS function $T_{lbs}$ applies the SMPL skeleton pose $p$ and blending weights $w$ to deform $x_c$ into observation space as $x_o$, where $w$ is provided by SMPL~\cite{SMPL:2015}. $Q$ here denotes the remaining parameters of the Gaussians, including scale $s$, opacity $\alpha$, and rotation $q$. For more details on canonical initialization, see Appendix A.1.1.

\subsubsection{Normal Map Rendering of Seen View} 
\label{frontNormalRendering}

We aim to faithfully capture the detailed surface geometry of dynamic human bodies from partial-view videos. Central to this process is the rendering of predicted normal maps, where the predicted \(\Delta n\) is applied to the initial SMPL normals \( n \) to compute \( n_c \) in canonical space. \( n_c \) is then transformed into the observation space \( n_o \) and rendered as normal maps \(I_n\). The Eq.~\eqref{eq:LBS}\&\eqref{eq:splatting} are modified for normals as:
\begin{equation}
  I_{n}=Splatting(x_o,n_o,Q,r),
  \label{eq:normalsplatting}
\end{equation}
\begin{equation}
  n_o = T_{lbs}(n_c,p,w).
  \label{eq:normalLBS}
\end{equation}
This transformation maps 3D Gaussians from the canonical space to the observation space, enabling the preservation of detailed geometry encoded by the normals and appearance.

We supervise the normal vectors using high-quality normal maps derived from ground truth RGB images. For this purpose, we leverage Sapiens~\cite{khirodkar2024sapiens} as a normal predictor to predict normal maps from video frames, using them as supervision for normal maps rendered in our observation space, expressed as:
% \begin{equation}
%   I_{pred} = Sapiens(I_{gt}),
%   \label{eq:normalGT}
% \end{equation}
\begin{equation}
\mathcal{L}_n = MSE(I_n,I^{\text{gt}}_{n}),
  \label{eq:normalloss}
\end{equation}
where $I^{\text{gt}}_{n}$ denotes the predicted normal map from Sapiens. The normal loss $\mathcal{L}_n$ is defined as the MSE loss between $I_{n}$ and $I^{\text{gt}}_{n}$. By aligning the predicted normal maps with those renderings, we achieve a high-fidelity representation of surface geometry that accurately captures both global and fine-grained details.

% \vspace{-3mm}
\subsection{Stage II: Invisible Appearance Reconstruction}
\label{fine-tuning}
Stage I produces an animatable 3D human model with visible appearances learned from partial-view video data, but the unobserved regions of the body typically suffer from relatively low visual quality. To ensure multi-view consistency for the unseen parts, we introduce a viewpoint-conditioned diffusion model as supervision, leveraging generative priors to predict the unseen views from the given inputs. Subsequently, we optimize the Gaussian decoder $G_{\theta}$ to reconstruct a fully renderable 3D human model from any viewpoint. To effectively utilize the observations and improve consistency between observed and hallucinated results, we introduce Dual-space Optimization, View Selection, and Pose Feature Injection techniques in the following.

\begin{figure}[tb]
    \centering
    \subfloat[]{\includegraphics[trim=30 50 710 0, clip,width=0.15\linewidth]{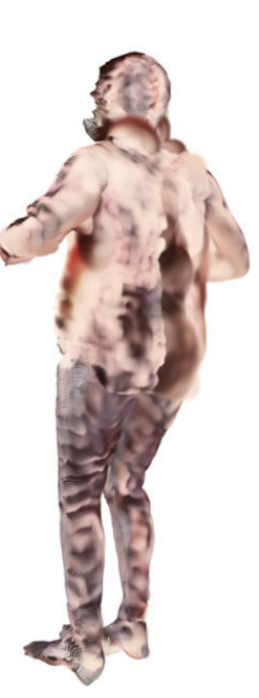}%
    \label{fig:ablation1-a}}
    \hfil
    \subfloat[]{\includegraphics[trim=170 50 570 0, clip,width=0.15\linewidth]{Figures/ablation_1.1.pdf}%
    \label{fig:ablation1-b}}
    \hfil
    \subfloat[]{\includegraphics[trim=300 50 440 0, clip,width=0.15\linewidth]{Figures/ablation_1.1.pdf}%
    \label{fig:ablation1-c}}
    \begin{tikzpicture}[overlay, remember picture]
        \draw[dotted, thick] (0, 3.6) -- (0, 0); % Adjust the coordinates based on your figure size
    \end{tikzpicture}
    \hfil
    \subfloat[]{\includegraphics[trim=450 50 290 0, clip,width=0.15\linewidth]{Figures/ablation_1.1.pdf}%
    \label{fig:ablation1-d}}
    \hfil
    \subfloat[]{\includegraphics[trim=730 50 10 0, clip,width=0.15\linewidth]{Figures/ablation_1.1.pdf}%
    \label{fig:ablation1-e}}
    \hfil
    \subfloat[]{\includegraphics[trim=590 50 152 0, clip,width=0.15\linewidth]{Figures/ablation_1.1.pdf}%
    \label{fig:ablation1-f}}

  \caption{\textbf{Left side: Dual-space Optimization} (a) w/o Dual-space Optimization; (b) w/ canonical optimization only; (c) w/ Dual-space Optimization; \textbf{Right side: Pose Feature Injection} (d) ground truth; (e) w/o pose feature injection; (f) w/ pose feature injection.}
 
  \label{fig:ablation1}
\end{figure}

\subsubsection{Dual-space Optimization}

Zero-1-to-3~\cite{liu2023zero1to3}, a viewpoint-conditioned diffusion model, is used to hallucinate full-body views from partial video frames, using reference frames from the monocular video and target view camera parameters as conditioning inputs.
Its explicit view control enables precise multi-view predictions for 3D reconstruction.  We leverage Score Distillation Sampling~(SDS)~\cite{poole2022dreamfusion} loss for predicting the unseen parts of our 3D Gaussian human model in the observation space. Unfortunately, naively combining Zero123 using SDS for dynamic human reconstruction leads to unrealistic reconstruction results. For instance, directly applying SDS in canonical space often results in quality degeneration in avatars—when generating 3D models with 2D diffusion models (See Fig.~\ref{fig:ablation1-b}).

To address this, we introduce Dual-space Optimization, which performs SDS optimization in both canonical and observation spaces. When conducting optimization in the canonical space, we use the rendering in the canonical space from Stage I as a conditioning reference for the 2D generative diffusion model. When conducting optimization in the observation space, we utilize the selected input images from the partial-view video as conditioning references. The SDS optimization process, combining Zero-1-to-3 with the Dual-space strategy, is thus expressed as:

\begin{equation}
  \mathcal{L}_{SDS}(I_{\theta})\triangleq \mathbb{E}_{t,\epsilon}[w(t)(\epsilon_{\phi}(z_t,y,R,T,t)-\epsilon)\frac{\partial I_{\theta}}{\partial\theta}],
  \label{eq:SDSObservOur}
\end{equation}
where $I_{\theta}$ represents a generated image from an unseen view in observation or canonical space.  $\epsilon_{\phi}$ is the predicted noise by Zero123 conditioned on the image $y$ and the target view camera parameters $(R, T)$. Since $\mathcal{L}_{SDS}$ is applied in both canonical and observation spaces, we take the observed frames from the input video as $y_{image}$ when optimizing in observation space, and take the canonical rendering from Stage I as $y_{image}$ when optimizing in canonical space.
This approach allows us to more effectively associate features across frames for the reconstruction of unseen parts.

During Dual-space Optimization, we found that appropriately balancing the training processes of the two spaces improves performance. As mentioned earlier, diffusion models face degeneration issues when optimizing in the canonical space. %Additionally, 
For instance, they struggle to predict accurate appearances for complex human poses in the observation space, often producing unrealistic 'tattoo-like' appearances, as shown in Fig.~\ref{fig:ablation_DS-c}. To address this, we set the weight between canonical and observation optimization as a hyperparameter in Stage II to enhance the overall process performance. This refined balance ensures better alignment of the model with the desired objectives, leading to more accurate and reliable outcomes in Stage II.

\subsubsection{View Selection}
\label{viewSelection}

The aforementioned Dual-space Optimization with SDS aids in synthesizing the unseen appearance of human avatars. Next, we introduce View selection to analyze which regions of the avatar are poorly observed.

\begin{figure}[tb]
    \centering
    \subfloat[]{\includegraphics[trim=25 50 30 20, clip,width=0.4\linewidth]{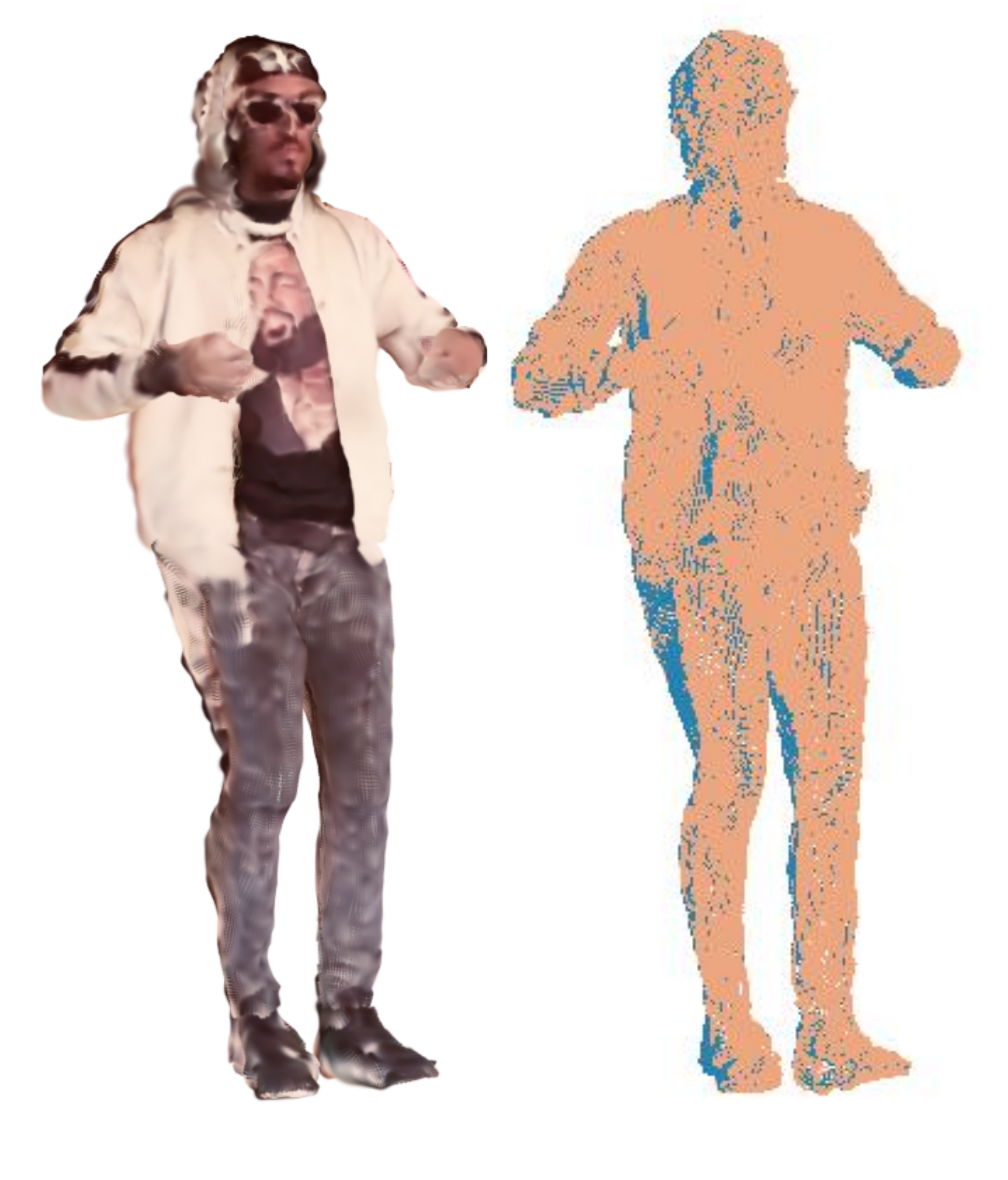}%
    \label{fig:visibility1}}
    \hfil
    \subfloat[]{\includegraphics[trim=70 70 10 50, clip,width=0.35\linewidth]{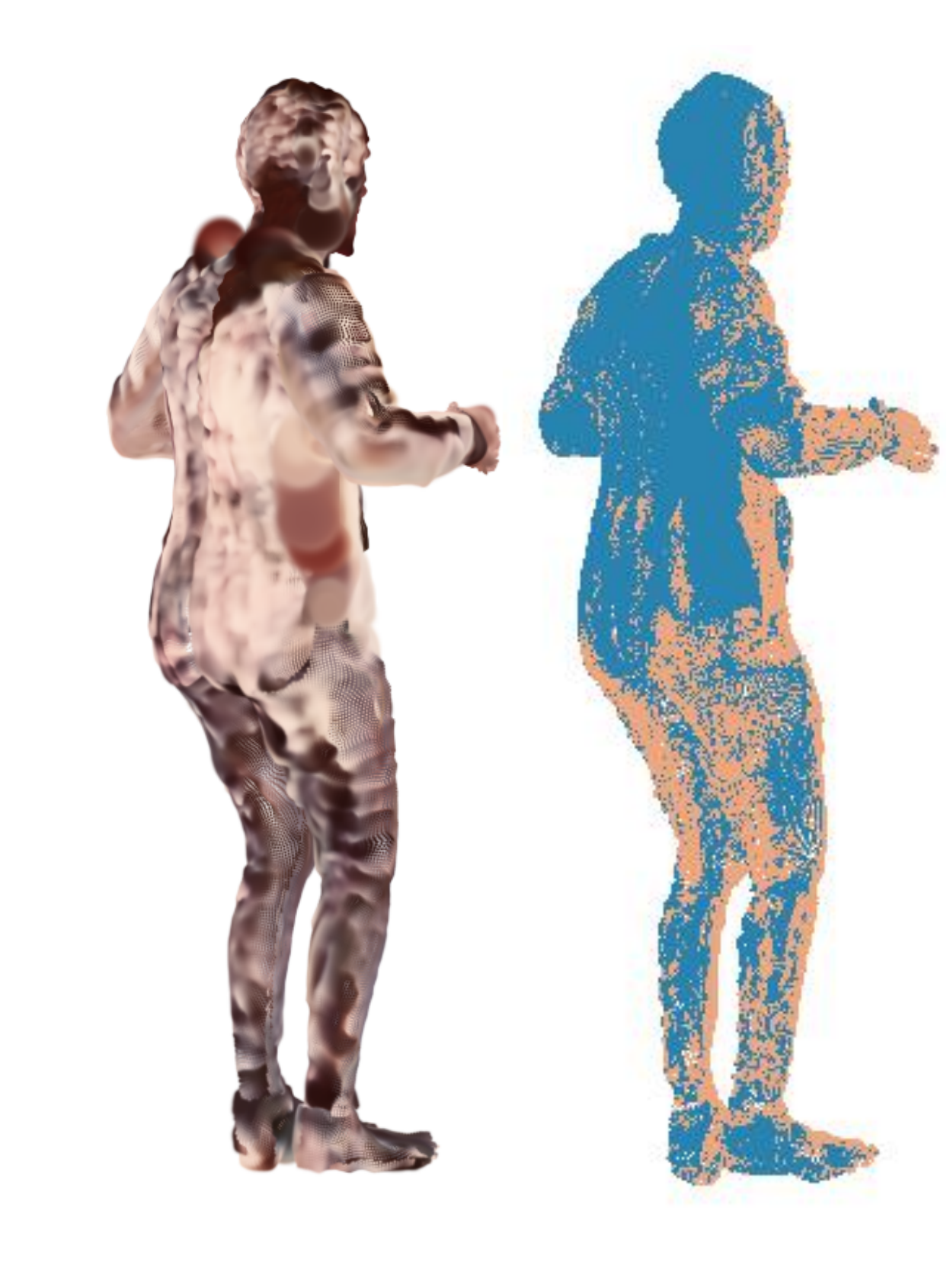}%
    \label{fig:visibility2}}
  \caption{\textbf{View Selection based on visibility map} (a) Seen view: Visible region (\textbf{\textcolor{myorange}{orange}}) covers more than 50\% of the foreground region; (b) Unseen view: Invisible region (\textbf{\textcolor{myblue}{blue}}) covers more than 50\% of the foreground region. }

  \label{fig:visibility}
\end{figure}

In both canonical and observation space optimizations, we identify the invisible views that require refinement. By utilizing the differentiable rasterization of Gaussian Splatting~\cite{kerbl3Dgaussians}, we determine the first intersecting Gaussian for each ray, marking these as visible points. Subsequently, visibility maps are rendered to differentiate between the visible and invisible regions of the human avatar as defined in Stage I. Specificly, we first estimate the visibility of each Gaussian. During the training of Stage I with seen views, given a ray $r$, the first Gaussian hit by the ray, $x$, is marked as a seen Gaussian, and its visibility $\psi$ is set to 1. Formally:
\begin{equation}
    \psi(x, r) = \begin{cases} 1, & x \text{ is the first Gaussian on } r \\ 0, & \text{otherwise} \end{cases},
    \label{eq:viewSelection1}
\end{equation}
where $\psi(x, r) = 1$ indicates that the Gaussian $x$ is visible in Stage I, while $\psi(x, r) = 0$ represents the opposite. 
After Stage I training is completed, a visibility map $I_v$ of a random viewpoint $v$ is rendered given $\psi(x, r)$, $r$, and the remaining attributes $Q$. In $I_v$, if the visible region $VR(I_v)$ covers less than 50\% of the foreground region $FR(I_v)$, this viewpoint is marked for refinement in Stage II. Then the View selection is expressed as:
\begin{equation}
    I_v = Splatting \left( x, \psi(x, r) , Q, r \right),
    \label{eq:viewSelection2}
\end{equation}
\begin{equation}
    Visibility(I_v) = \frac{VR(I_v)}{FR(I_v)},
    \label{eq:viewSelection3}
\end{equation}
\begin{equation}
    \mathbb{I}_v = \begin{cases} 0, & Visibility(I_v) \leq 50\% \\ 1, & \text{otherwise} \end{cases}, 
    \label{eq:viewSelection4}
\end{equation} 
where $\mathbb{I}_v = 0$ signifies an unseen viewpoint, indicating that the avatar needs to be refined from viewpoint $v$ in Stage II. And visibility map examples are shown in Fig.~\ref{fig:visibility}.

\subsubsection{Pose Feature Injection}
\label{poseInject}
Furthermore, during Dual-space Optimization, while SDS is applied across diverse poses in observation spaces, the Gaussian decoder is trained in the canonical space. To capture pose-dependent appearances in the observation space, such as garment wrinkles in Fig.~\ref{fig:ablation1-d}, we leverage the pose encoder similar to GaussianAvatar~\cite{hu2023gaussianavatar} to extract pose-related features, which are then injected into the decoder network. Consequently,  we have:
\begin{equation}
  (\Delta x, \Delta n,c,s)=G_{\theta}([S,P]),
  \label{eq:stage2G_decoder}
\end{equation}
\begin{equation}
  P = Encoder(P_{uv}),
  \label{eq:poseencoder}
\end{equation}
where $P_{uv}$ is the UV positional map of SMPL for each pose, and $P$ denotes the extracted pose feature, which is concatenated with the canonical features $S$ as input of the Gaussian decoder $G_{\theta}$. And $Encoder(.)$ maps $P_{uv}$ to $P$. All the outputs of $G_{\theta}$ remain the same as in Stage I.

\subsubsection{Normal Map Supervision of Unseen View} 
\label{BackNormalRendering}
For the reconstruction of unseen-view geometry, we extend the rendering process described in Sec.~\ref{frontNormalRendering} to generate normal maps for unseen views. Specifically, the normal maps of given views are treated as front normal maps. To compute the back normal maps from their corresponding front normal map, we utilize a depth-aware, silhouette-consistent bilateral normal integration (d-BiNI) method~\cite{bini2022cao}. These back normal maps are then combined with pretrained SMPL-aware IF-Nets~\cite{chibane20ifnet}, which inpaint the geometry of the remaining body regions. The resulting output is a complete set of normal maps, which serves as full-body normal supervision in Eq.~\ref{eq:normalloss}.

\subsection{Training Losses}
% \subsection{Optimization}
\label{trainingLosses}
In Stage I, we are modeling a dynamic avatar from partial-view videos using a Gaussian decoder. Additionally, we refine the input pose to correct inaccuracies from SMPL fitting. This stage utilizes mean squared error(MSE) loss, structural similarity(SSIM) loss~\cite{SSIM}, and perceptual similarity(LPIPS) loss~\cite{zhang2018unreasonable} between the predicted RGB images and ground truth, as $\mathcal{L}_{rgb}$, $\mathcal{L}_{ssim}$, and $\mathcal{L}_{lpips}$, respectively.
We also apply Frobenius Norm loss as regularization term for optimizable canonical features $S$, offset $\Delta x$, and scale $s$:
\begin{equation}
\begin{aligned}
\mathcal{L}_{f}^{S} = \sqrt{\sum_{k=1}^{n} |S_{k}|^{2}}, 
\mathcal{L}_{f}^{\Delta x} = \sqrt{\sum_{k=1}^{n} |\Delta x_{k}|^{2}},
\mathcal{L}_{f}^{s} = \sqrt{\sum_{k=1}^{n} |s_{k}|^{2}},
\end{aligned}
\label{eq:lossL2}
\end{equation}
where $\mathcal{L}_{f}^{S}, \mathcal{L}_{f}^{\Delta x},\mathcal{L}_{f}^{s}$ denotes the loss of the $S$, $\Delta x$, and $s$, respectively.
Combining with the normal loss $\mathcal{L}_{n}$ from Eq.~\ref{eq:normalloss}, the total loss function for Stage I is as follows:
\begin{equation}
\begin{aligned}
\mathcal{L}_{StageI}= &\lambda_{rgb}\mathcal{L}_{rgb}+\lambda_{n}\mathcal{L}_{n}+\lambda_{ssim}\mathcal{L}_{ssim}\\
&+\lambda_{lpips}\mathcal{L}_{lpips}+\lambda_{\Delta x}\mathcal{L}_{f}^{\Delta x}+\lambda_{s}\mathcal{L}_{f}^{s}+\lambda_{S}\mathcal{L}_{f}^{S}.
\end{aligned}
  \label{eq:lossstage1}
\end{equation}

In Stage II, the pose encoder and Gaussian decoder are optimized using SDS losses. To prevent degradation of visible appearance and geometry, $\mathcal{L}_{StageI}$ is incorporated. Additionally, $\mathcal{L}_{f}^{p}$ is added with Frobenius Norm loss to regularize the pose feature map. The total loss function for Stage II is expressed as:
\begin{equation}
\begin{aligned}
\mathcal{L}_{StageII}=&\mathcal{L}_{StageI}+\lambda_{p}\mathcal{L}_{f}^{p}\\
&+\lambda_{SDS}(\mathcal{L}_{SDS}^{o}+\mathcal{L}_{SDS}^{c}),
\end{aligned}
  \label{eq:lossstage2}
\end{equation}
where $\mathcal{L}_{SDS}^{o}$ and $\mathcal{L}_{SDS}^{c}$ denote the SDS loss in observation and canonical space, respectively, as defined in Eq.~(\ref{eq:SDSObservOur}).

Furthermore, we design a progressive training strategy in this stage, gradually diminishing the weight of SDS loss. This strategy is employed to further enhance the effectiveness and efficiency of the visible appearance reconstruction. More details on progressive training are in Appendix A.2.2.

\newcommand{\centered}[1]{\begin{tabular}{l} #1 \end{tabular}}

\begin{figure*}[!t]
\centering
\includegraphics[width=1\textwidth]{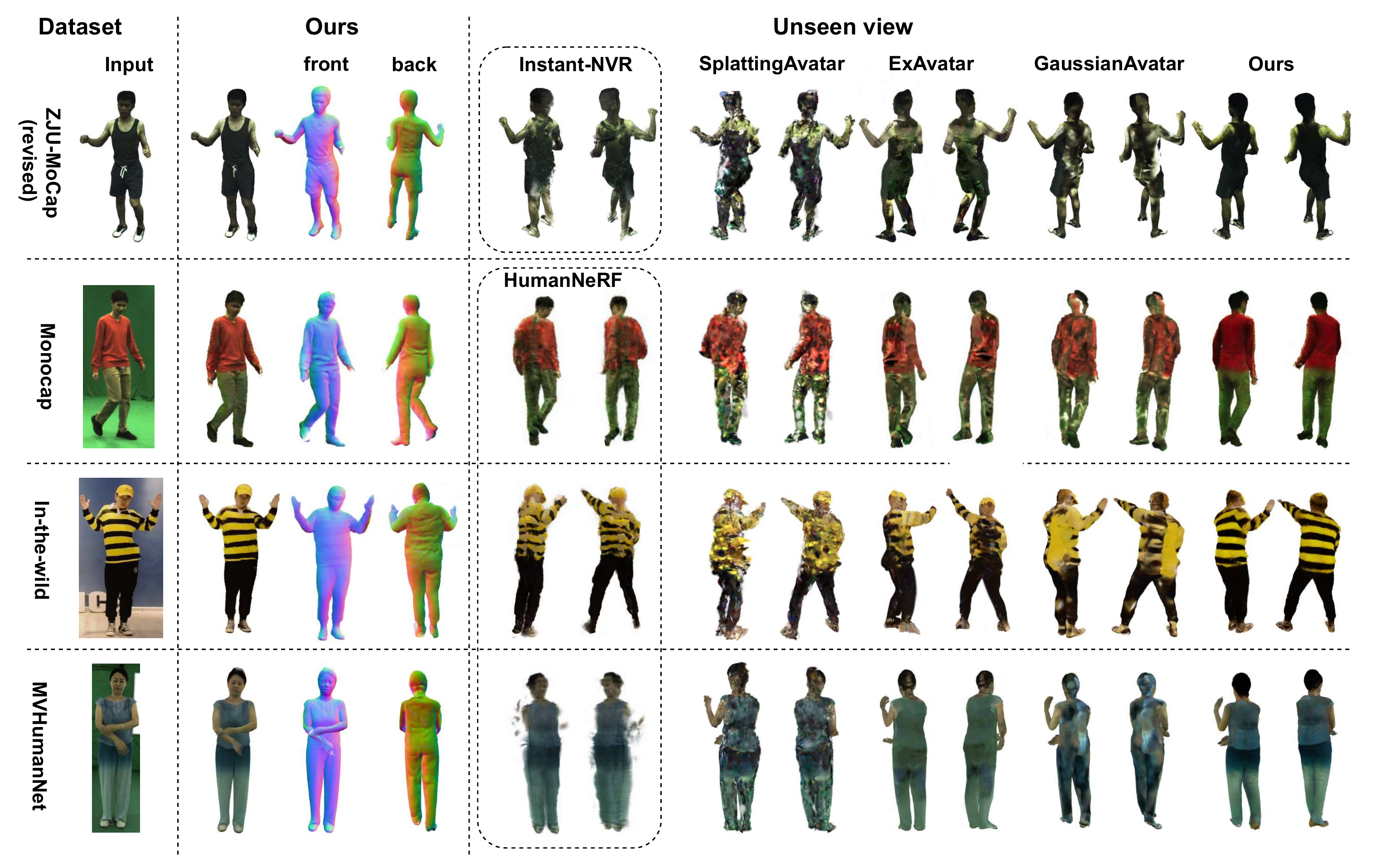} 
\caption{Qualitative comparison on four datasets. We compare the novel view synthesis quality with HumanNeRF~\cite{weng2022humannerf}, Instant-NVR~\cite{instant_nvr},  SplattingAvatar~\cite{shao2024splattingavatar}, ExAvatar~\cite{moon2024exavatar} and GaussianAvatar~\cite{hu2023gaussianavatar}.}
\label{fig: qualitative1}
\vspace{-3mm}
\end{figure*}

\section{Experiments}

\subsection{Datasets}

\hspace*{1.5em}\textbf{ZJU-Mocap(revised) dataset~\cite{peng2021neural}.} This dataset is a multi-view dataset. We train and test using this dataset following Instant-NVR~\cite{instant_nvr}. One specific camera view is used as the monocular training input, while six cameras evenly distributed around the object are reserved for comprehensive evaluation.

\textbf{Monocap dataset.} Similar to ZJU-Mocap(revised), the Monocap dataset contains multi-view videos collected by AnimatableNeRF~\cite{peng2024animatable} from the DeepCap dataset~\cite{habermann2020deepcap} and the DynaCap dataset~\cite{habermann2021}. The dataset setting follows the ZJU-Mocap(revised) dataset.

\textbf{MVHumanNet dataset~\cite{xiong2024mvhumannet}.} The dataset is a large-scale collection of multi-view human images, encompassing human masks, camera parameters, 2D and 3D keypoints, SMPL/SMPLX parameters. The dataset setting follows the ZJU-Mocap(revised) dataset as well.

\textbf{In-the-wild dataset.} This dataset contains YouTube videos collected by HumanNeRF~\cite{weng2022humannerf} and Dance Dance Generation~\cite{zhou2019dance}. We employ BEV~\cite{BEV} to estimate camera parameters and the SMPL bodies, then utilize Xmem~\cite{cheng2022xmem} along with Segment-anything~\cite{kirillov2023segany} to extract foreground segmentation of video frames.

\begin{figure*}[!t]
\centering
\includegraphics[width=\textwidth]{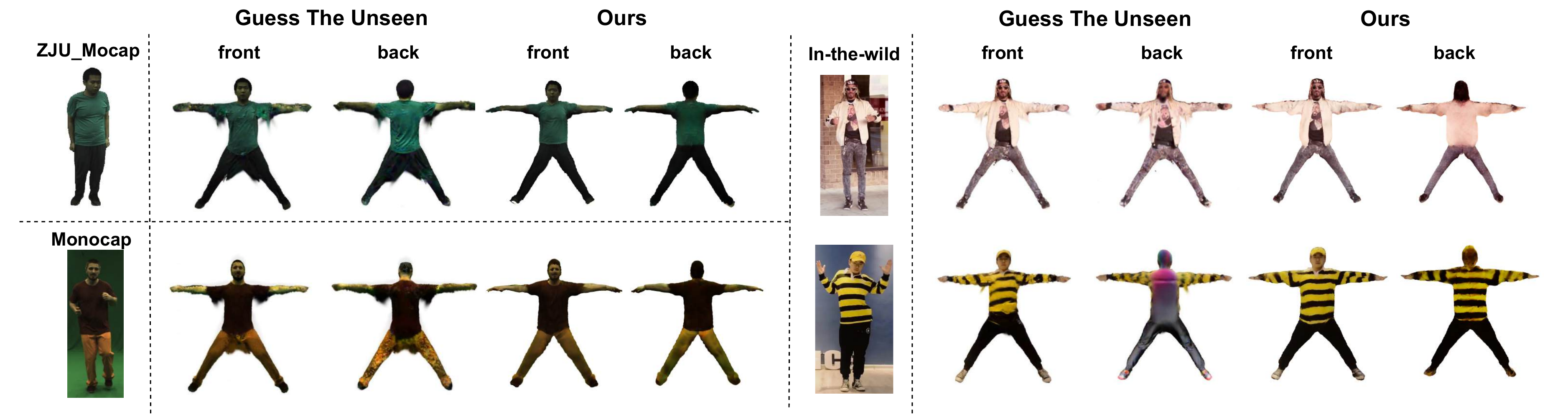} % Reduce the figure size so that it is slightly narrower than the column.

\caption{Qualitative comparison on three datasets. We compare the novel view synthesis quality with GuessTheUnseen~\cite{lee2024gtu}.}
\label{fig: qualitative2}

\end{figure*}

\begin{table*}[!t]
 
\centering
\renewcommand{\arraystretch}{1.5}
\begin{tabular}{c|ccc|ccc|ccc}
\hline
\multirow{2}{*}{} & \multicolumn{3}{c|}{ZJU-Mocap(revised)} & \multicolumn{3}{c|}{Monocap} & \multicolumn{3}{c}{MVHumanNets} \\ 
                  & PSNR$\uparrow$ & SSIM$\uparrow$ & LPIPS$\downarrow$ & PSNR$\uparrow$ & SSIM$\uparrow$ & LPIPS$\downarrow$ & PSNR$\uparrow$ & SSIM$\uparrow$ & LPIPS$\downarrow$ \\ \hline
HumanNeRF         &---&---&---&19.21&0.9456&0.0715&19.38&0.9416&0.0706\\ 
Instant-NVR       &19.90&0.9458&0.0630 &20.03&0.9469&0.0613 &---&---&---\\
SplattingAvatar   &19.37&0.9436&0.0702 &19.51&0.9444&0.0697 &19.49&0.9467&0.0689 \\
ExAvatar          &19.65&0.9449&0.0678 &19.65&0.9475&0.0676 &19.69&0.9470&0.0671 \\
GaussianAvatar    &19.50&0.9434&0.0687 &19.67&0.9489&0.0635 &19.70&0.9471&0.0645 \\
GuessTheUnseen    &20.06&0.9493&0.0615 &20.56&0.9502&0.0598 &---&---&---         \\
 \textbf{Ours}    &\textbf{20.82}&\textbf{0.9552}&\textbf{0.0569}   & \textbf{21.16}& \textbf{0.9532}& \textbf{0.0549} &\textbf{20.98}&\textbf{0.9517}&\textbf{0.0553}\\ \hline
\end{tabular}
\caption{Quantitative evaluation on ZJU-Mocap(revised), MVHumanNet, and Monocap datasets (unseen view only).\label{Q_table}}

\end{table*}

\subsection{Implementation Details} 
\label{implementaion}
All videos from all datasets are clipped to 3-5 seconds (100-150 frames) and exclusively capture front views of the subjects. During stage I, training is conducted on a single RTX-3090 GPU with a batch size of 2, requiring approximately 1 hour for 200 training epochs. In Stage II, the entire framework is trained on two RTX-3090 GPUs with a batch size of 1, while the diffusion model is loaded exclusively on the second GPU. Overall, the Gaussian parameter decoder consists of approximately 186K parameters. The training process requires approximately 28 GB of GPU memory. Training requires approximately 2–3 hours for 400 epochs, while inference operates at a speed of 13–18 frames per second, depending on the resolution.

\subsection{Comparisons with Video-based Methods}

We conduct comparisons of our method with HumanNeRF~\cite{weng2022humannerf}, Instant-NVR~\cite{instant_nvr}, SplattingAvatar~\cite{shao2024splattingavatar}, ExAvatar~\cite{moon2024exavatar}, GaussianAvatar~\cite{hu2023gaussianavatar}, and GuessTheUnseen~\cite{lee2024gtu}.

For a fair comparison, Instant-NVR~\cite{instant_nvr} is trained on the revised version of the ZJU-Mocap dataset, which offers refined camera parameters, SMPL fittings, and more accurate instance masks with body-part segmentation, crucial for the execution of their method. However, HumanNeRF is not adapted to this dataset, and MVHumanNet and Monocap are applied to evaluate this method. Additionally, Instant-NVR lacks a pose refinement technique akin to HumanNeRF, which assists in addressing inaccurate fitting issues in the in-the-wild dataset. GuessTheUnseen is evaluated on the ZJU-Mocap(revised) and Monocap datasets but not on MVHumanNet, as the original images in this dataset contain black bounding boxes that significantly hinder the human motion detection process performed by GuessTheUnseen. Therefore, we will discuss the comparison results based on the type of dataset utilized.

\subsubsection{ZJU-Mocap(revised), MVHumanNet, and Monocap Datasets}
These three datasets serve as the primary testbeds for our experiments due to the availability of ground truths for invisible parts. The quantitative results are presented in Tab.~\ref{Q_table}, where our method surpasses all evaluation techniques in all three metrics, indicating its efficacy in reconstructing both geometry and appearance for invisible parts. Qualitatively, as shown in Fig.~\ref{fig: qualitative1}, the limitations of all compared methods become more evident when visualizing invisible parts. Methods such as GaussianAvatar, ExAvatar, and SplattingAvatar, which are based on Gaussian Splatting, exhibit noticeable artifacts and inconsistencies, including noisy textures and blank spots. Instant-NVR and HumanNeRF, due to their NeRF-based ray-shooting geometric reconstruction technique, not only struggle with appearance consistency but also suffer from geometric issues like floating artifacts and penetrating holes, diminishing the realism of the synthesized avatars. While GuessTheUnseen can infer unseen-view appearances, it introduces noisy textures and multi-face `Janus' artifacts, as shown in Fig.~\ref{fig: qualitative2}.

\subsubsection{In-the-wild Dataset} 

The in-the-wild dataset comprises various monocular dancing videos sourced from the internet. Due to the lack of novel view references, our primary focus is on qualitative evaluation results in comparison to other methods for the unseen parts of humans. As shown in Fig.~\ref{fig: qualitative1}\&\ref{fig: qualitative2}, we observe that HumanNeRF faces similar challenges to Instant-NVR. The generative networks struggle to effectively fuse sampling points for novel view rendering due to a lack of supervision. This shortcoming results in floating points and ``foggy" artifacts in the rendered outputs. Additionally, the Gaussian-based methods continue to produce unrealistic ``tattoo-like" appearances on the backs of synthesized avatars, highlighting their limitations in preserving overall appearance fidelity. For GuessTheUnseen, the `Janus' artifacts become more pronounced, and it even fails to correctly infer the appearance of some subjects. In contrast, our method consistently demonstrates superior performance in addressing the challenges of invisible parts synthesis, excelling in both geometry and appearance reconstruction. The invisible parts of the synthesized avatars show not only enhanced geometric precision but also significantly improved appearance fidelity, with fewer artifacts and smoother textures.

\begin{figure*}[!t]
\centering
\includegraphics[width=\textwidth]{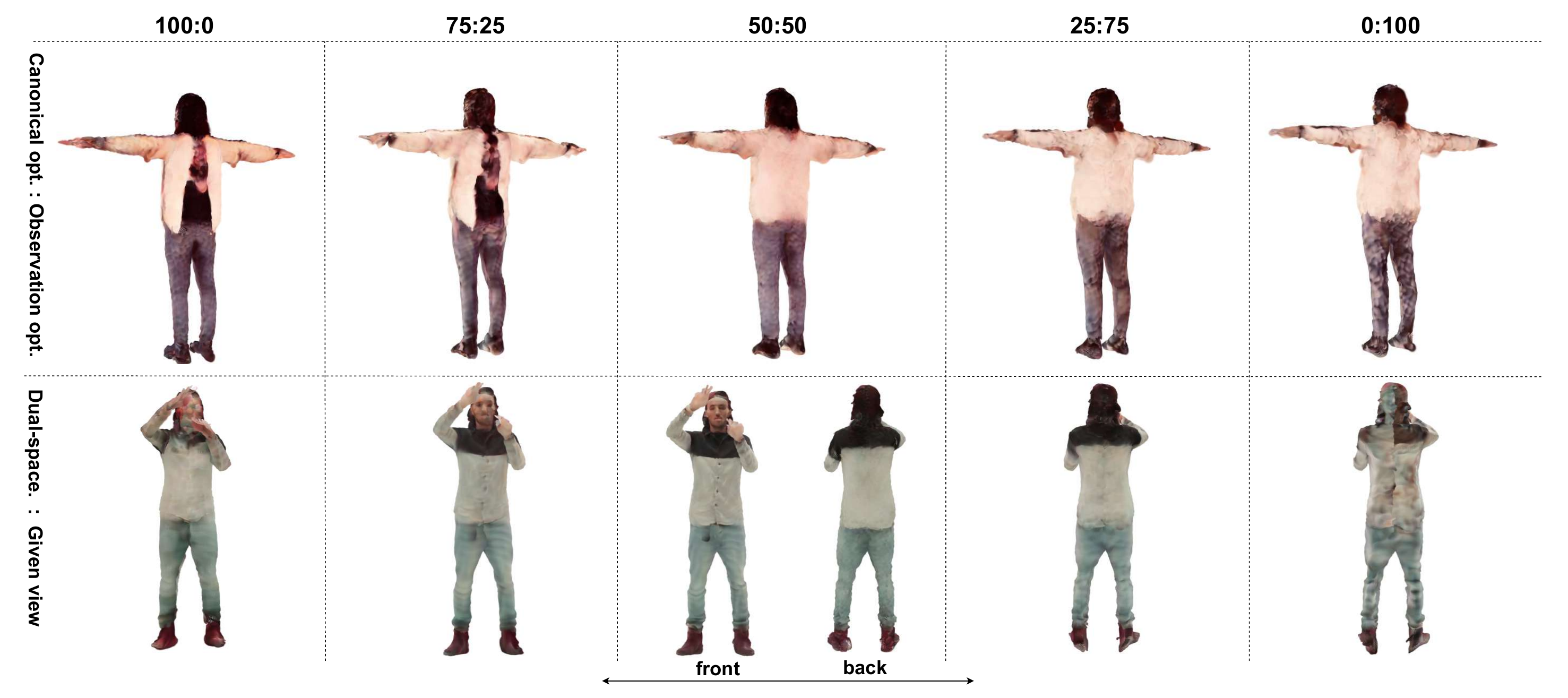} % Reduce the figure size so that it is slightly narrower than the column.

\caption{\textbf{First row:} Training ratio between canonical optimization and observation optimization(Canonical opt. : Observation opt.). \textbf{Second row:} Training ratio between Dual-space Optimization and given-view training(Dual-space : Given view).}
\label{fig: percentage_training}

\end{figure*}

\begin{table}[]

    \centering
     \renewcommand{\arraystretch}{1.5}
    \begin{tabular}{l|l|l|l}
    \hline
         & PSNR$\uparrow$& SSIM$\uparrow$ &LPIPS$\downarrow$ \\ \hline
        full model  & \textbf{21.06} &\textbf{0.9536} & \textbf{0.0551}\\ \hline
         full model w/o Prog. & 20.98 & 0.9523 & 0.0559\\ \hline
         full model w/o Canonical. & 20.16 & 0.9503 & 0.0586\\ \hline
        full model w/o View-selection. & 20.03 & 0.9489 & 0.0602\\ \hline
        full model w/o Pose-injection. & 19.78 & 0.9473 & 0.0626\\ \hline
       full model w/o Observ. & 19.56 & 0.9434 & 0.0685\\ \hline
    \end{tabular}
 \caption{Quantitative results for ablation study. Prog. is short for progressive training strategy.}\label{tab:ablation}
\vspace{-5mm}
\end{table}

\subsection{Ablation Study}
\label{}
\subsubsection{Dual-space Optimization} 
\begin{figure}[]
    \centering
    \subfloat[]{\includegraphics[trim=20 75 18 70, clip,width=0.20\linewidth]{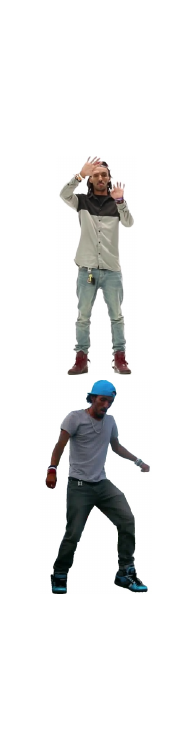}%
    \label{fig:ablation_DS-a}}
    \hfil
    \subfloat[]{\includegraphics[trim=24 60 684 25, clip,width=0.21\linewidth]{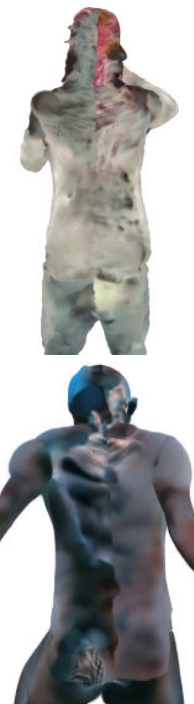}%
    \label{fig:ablation_DS-b}}
    \hfil
    \subfloat[]{\includegraphics[trim=121 60 587 25, clip,width=0.21\linewidth]{Figures/ablation_2.1.pdf}%
    \label{fig:ablation_DS-c}}
    \hfil
    \subfloat[]{\includegraphics[trim=214 60 493 25, clip,width=0.21\linewidth]{Figures/ablation_2.1.pdf}%
    \label{fig:ablation_DS-d}}
  
  \caption{\textbf{Ablation study about Dual-space Optimization} (a) w/o Dual space Optimization. (b) w/ observation optimization only. (c) full model novel view.}
  
  \label{fig:ablation_DS}
\end{figure}

Next, we evaluate the effectiveness of Dual-Space Optimization, with ablation results presented in Fig.~\ref{fig:ablation_DS}. Observation optimization is crucial for reconstructing the invisible parts of the avatar, but it often encounters challenges and may not converge effectively, resulting in rough and less satisfactory appearances. In such cases, the canonical optimization step becomes essential. Leveraging the canonical space, the optimization converges more effectively, yielding smoother and more visually pleasing results. Nevertheless, observation optimization can mitigate the quality degradation issues that arise when relying solely on SDS optimization in the canonical space, as shown in Fig.~\ref{fig:ablation1-a},\ref{fig:ablation1-b}\&\ref{fig:ablation1-c}. This iterative approach highlights the importance of both observation and canonical optimization for achieving optimal results.

The balance between these two optimizations is crucial. As shown in Fig.~\ref{fig: percentage_training} and Tab.~\ref{tab: ablation_1}, the optimal ratio between canonical-space optimization and observation-space optimization is 50:50 within each training epoch. A higher proportion of canonical optimization leads to a double-face artifact, while a higher proportion of observation-space optimization results in degraded quality in the back-view appearance. 

Furthermore, we continue given-view training in Stage II to preserve sharp front-view quality. As shown in Fig.~\ref{fig: percentage_training} and Tab.~\ref{tab: ablation_1}, a higher ratio of Dual-space Optimization leads to lower reconstruction quality and fidelity in the front view of the avatar. However, when the Dual-space Optimization ratio drops below 50\%, the quality of the predicted back-view appearance degrades. Therefore, we empirically adopt a 50:50 training ratio to balance the quality of both views.
\begin{table}[]
 
\centering
\renewcommand{\arraystretch}{1.5}
\begin{tabular}{l|ccc|ccc}
\hline
\multirow{2}{*}{} & \multicolumn{3}{c|}{Dual-space. : Given view} & \multicolumn{3}{c}{Canonical : Observation}  \\ \hline
                  & PSNR$\uparrow$ & SSIM$\uparrow$ & LPIPS$\downarrow$ & PSNR$\uparrow$ & SSIM$\uparrow$ & LPIPS$\downarrow$  \\ \hline
100 : 0     &19.89&0.9459&0.0626&19.53&0.9456&0.0715\\ 
 75 : 25    &20.56&0.9496&0.0593&19.93&0.9463&0.0698\\
 50 : 50    &\textbf{21.06}&\textbf{0.9536}&\textbf{0.0551}&\textbf{21.06}&\textbf{0.9536}&\textbf{0.0551}\\
 25 : 75    &20.67&0.9503&0.0586&20.69&0.9490&0.0596\\
  0 : 100   &19.23&0.9436&0.0667&20.06&0.9471&0.0629\\ \hline
\end{tabular}
\caption{Quantitative evaluation on training percentage ablation: Dual-space Optimization: Given view training and Canonical optimization : Observation optimization.\label{tab: ablation_1}}

\end{table}

\begin{table}[]
    \centering
     \renewcommand{\arraystretch}{1.5}
    \begin{tabular}{l|ccc}
    \hline
         & PSNR$\uparrow$& SSIM$\uparrow$ &LPIPS$\downarrow$ \\ \hline
        0\%  & 19.80 &0.9463 &0.0634\\ 
         25\% & 20.89 & 0.9516 & 0.0549\\ 
         50\% &\textbf{21.06}&\textbf{0.9536}&\textbf{0.0551}\\ 
        75\% & 20.76 & 0.9503 & 0.0593\\ 
        100\% & 20.16 & 0.9493 & 0.0601\\ \hline
      
    \end{tabular}
 \caption{Quantitative evaluation on visibility ratios.}\label{tab:ablation_2}

\end{table}

\begin{figure*}[tb]
\centering
\includegraphics[width=\textwidth]{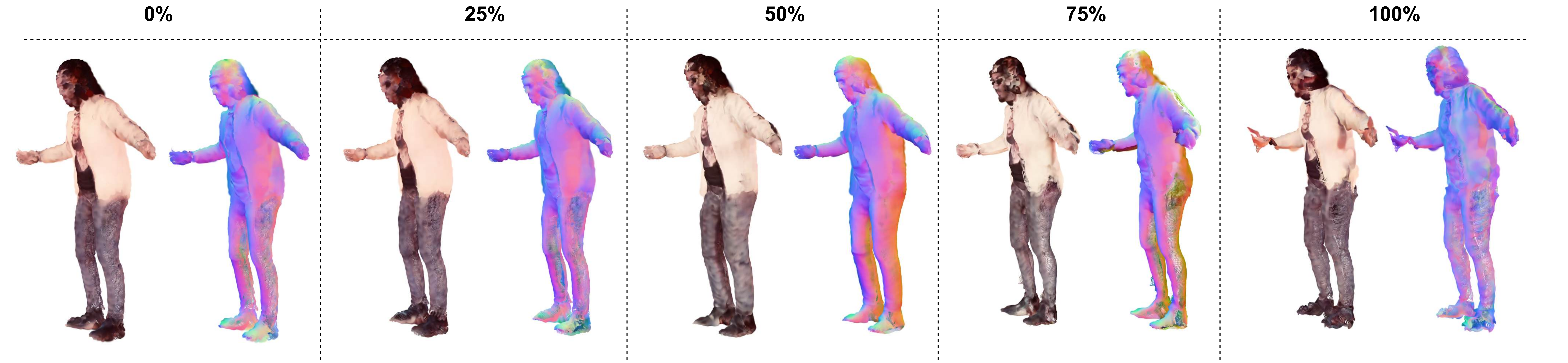} % Reduce the figure size so that it is slightly narrower than the column.

\caption{Qualitative comparison of results under varying visibility ratios.} 

\label{fig: view_selection_ratio}

\end{figure*}

\begin{figure}[!t]
    \centering   
    \subfloat[]{\includegraphics[trim=20 0 360 20, clip,width=0.23\linewidth]{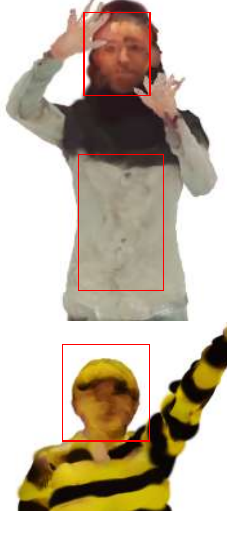}%
    \label{fig:ablation_V&P-a}}
    \hfil
    \subfloat[]{\includegraphics[trim=136 0 245 20, clip,width=0.23\linewidth]{Figures/ablation_2.3.pdf}%
    \label{fig:ablation_V&P-b}}
    \hfil
    \subfloat[]{\includegraphics[trim=253 0 130 20, clip,width=0.23\linewidth]{Figures/ablation_2.3.pdf}%
    \label{fig:ablation_V&P-c}}
     \subfloat[]{\includegraphics[trim=360 0 24 20, clip,width=0.23\linewidth]{Figures/ablation_2.3.pdf}%
    \label{fig:ablation_V&P-d}}

  \caption{\textbf{Ablation study about View Selection (View.) and Pose Feature Injection (Pose.) } (a) w/o both. (b) w/ View. and w/o Pose. (b) w/o View. and w/ Pose. (d) full model.}
  \label{fig:ablation_V&P}

\end{figure}

\subsubsection{View Selection and Pose Feature Injection} 
We investigate the influence of View Selection and Pose Feature Injection in the following. As shown in Fig.~\ref{fig:ablation_V&P}, View selection filters out visible views, preserving the alignment between visible and canonical appearances, thereby reducing potential disruptions from observation optimization. Additionally, pose feature injection plays a crucial role in further enhancing dynamic appearance, allowing for the capture of finer details, especially in facial regions and cloth textures. These improvements significantly contribute to the overall fidelity and realism of the synthesized avatars. In Tab.~\ref{tab:ablation}, we present a quantitative evaluation of all components in the ZJU-Mocap(revised) dataset, MVHumanNet dataset, and Monocap dataset. Our full model with progressive training achieves the best results in this evaluation.

In practice, we set the visibility ratio to 50\% in Eq.~\ref{eq:viewSelection4} to determine whether a surface region is sufficiently observed during Stage I. This threshold is critical for balancing the learning focus between well-observed and less-visible regions in Stage II. As shown in Fig.~\ref{fig: view_selection_ratio}, when the ratio is too low, Dual-space Optimization overfits front-view regions, producing low-fidelity artifacts and limited generalization. Conversely, setting it too high shifts the focus to poorly observed regions (e.g., side views), leading to incomplete supervision and degraded quality. Tab.~\ref{tab:ablation_2} reports that a 50\% ratio provides a good trade-off, preserving high-quality reconstruction for both front and side views.

\section{Discussion and Conclusion}
\subsection{Limitation}
\begin{figure}[tb]
\centering
\includegraphics[width=\linewidth]{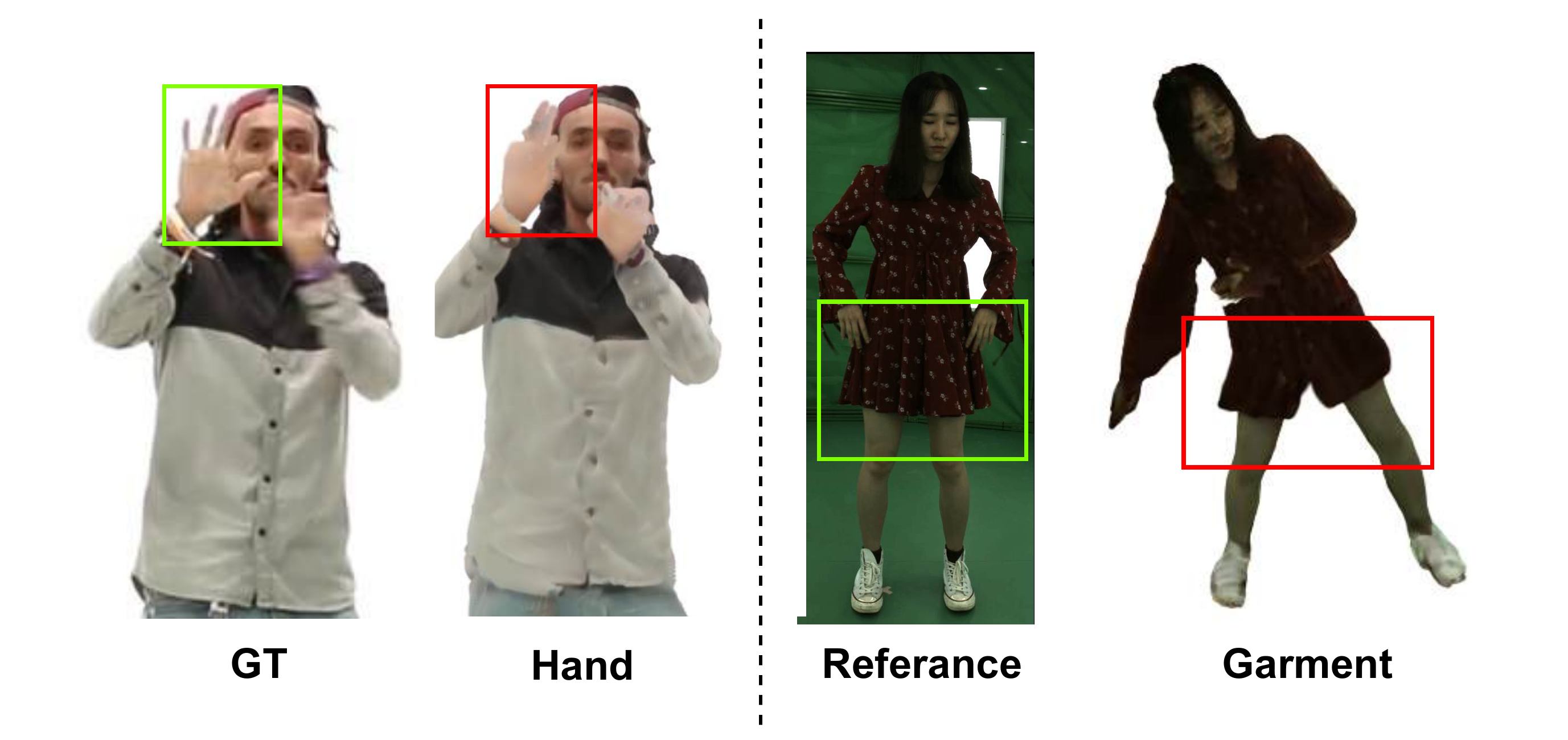} % Reduce the figure size so that it is slightly narrower than the column.

\caption{Failure cases on hand and garments.}
\vspace{-5mm}
\label{fig: F_cases}

\end{figure}

Since our method depends on human body fitting and foreground segmentation, artifacts may occur due to inaccuracies in these videos within processes. Despite incorporating pose optimization to correct poses, the reconstruction of hand parts and body shape may still exhibit artifacts in certain cases, as shown on the left side of Fig.~\ref{fig: F_cases}. While our approach generally yields more realistic results, similar to many existing methods~\cite{instant_nvr,weng2022humannerf,hu2023gaussianavatar}, it still faces challenges in accurately modeling loose attire, such as dresses. This limitation, illustrated on the right side of Fig.~\ref{fig: F_cases}, underscores the need for future enhancements to better capture complex clothing geometry and dynamics.

\subsection{Conclusion}

In this paper, we introduce \name, a novel approach for high-quality dynamic human reconstruction from monocular videos. By leveraging 2D diffusion model priors, \name effectively reconstructs and infers the unseen parts of 3D human avatars. We introduce Dual-Space Optimization, which applies Score Distillation Sampling (SDS) in both canonical and observation spaces, ensuring visual consistency and realism across various poses. Furthermore, View Selection and Pose Feature Injection strategies resolve conflicts between SDS predictions and observed data, enhancing overall avatar fidelity. Extensive experiments on benchmarks demonstrate that \name outperforms state-of-the-art methods, particularly in rendering the unseen parts of the human body.

\section*{Acknowledgment}
The authors would like to thank the anonymous reviewers for their constructive comments and suggestions. X. Dong is supported in part by Guangdong and Hong Kong Universities "1+1+1" Joint Research Collaboration Scheme.The opinions expressed are solely those of the authors, and do not necessarily represent those of the funding agencies.

\appendices
\section{Training Details}
\label{sec:details}
This section provides more details about the implementation and training of our method. The entire pipeline was evaluated on a system equipped with two RTX 3090 GPUs using a 1920×1080 resolution video. The preprocessing stage, which includes SMPL~\cite{SMPL:2015} human body fitting and Sapiens~\cite{khirodkar2024sapiens} normal prediction, takes approximately 40 minutes in total (20 minutes each). This is followed by Stage I training, which requires around 40 minutes, and a subsequent post-processing step to generate normal maps for unseen views, taking an additional 20 minutes. Finally, Stage II training takes approximately 3 hours, bringing the total processing time to just over 4.5 hours.

\subsection{Stage I}

\subsubsection{Canonical Initialization}
We unwrap the T-pose body onto a UV map, where each pixel stores a 3D position vector. The positional UV map, with a resolution of $(512 \times 512 \times 3)$, is used to initialize Gaussians in the canonical space, ensuring proper alignment with the body’s structure. Additionally, a downsampled $(128 \times 128 \times 3)$ version of the positional UV map serves as input to the Gaussian decoder, aiding in reconstructing and refining the 3D representation.

\subsubsection{Model Architecture}
The Gaussian parameter decoder consists of an 8-layer Multi-Layer Perceptron (MLP) with intermediate layer dimensions of $(128\times128, 128\times128, 128\times128, 256\times256, 128\times128, 128\times128, 128\times128, 64\times64)$, and includes a skip connection from the input to the fourth layer. It is followed by four separate prediction heads, each implemented as a 2-layer MLP, responsible for predicting the Gaussian position offsets $\Delta x$, normal offsets $\Delta n$, colors $c$, and scales $s$, respectively.

\subsubsection{Training}
The training objectives in this stage focus on image losses and optimizations about Gaussian parameters. We set weights for each objective as $\lambda_{rgb} = 0.8$, $\lambda_{n} = 0.8$, $\lambda_{ssim} = 0.2$, $\lambda_{lpips} = 0.2$, $\lambda_{\Delta x} = 0.85$, $\lambda_{s} = 0.03$, $\lambda_{S} = 1$. 

\subsubsection{Pose Optimization} Our method leverages pose optimization from GaussianAvatar~\cite{hu2023gaussianavatar} for the In-the-wild dataset as a correction for fitted SMPL~\cite{SMPL:2015} pose parameters. We have omitted this functionality for the ZJU-Mocap dataset, as their ground truth pose is accurate. However, GaussianAvatar keeps optimizing pose parameters for the ZJU-Mocap dataset, which leads to inaccurate poses, especially for invisible parts. Please check the accompanying video results for more details. 
\subsection{Stage II} 
\subsubsection{Dual-space Optimization}
\label{training_dual_space}
In this stage, we apply Dual-space Optimization on top of visible appearance reconstruction to predict the invisible appearance. During training, each epoch is divided into two parts: 50\% for given view training and 50\% for Dual-space Optimization. In Dual-space Optimization, the weight of canonical optimization is treated as a hyperparameter, defaulting to 50\%. In practice, ratios between optimizations around 50:50 (e.g., 40:60 or 60:40) also yield acceptable results depending on the subject. In this work, we empirically adopt a 50:50 split as our default setting. The fine-tuning losses are added on top of $\mathcal{L}_{StageI}$. We set $\lambda_{p} = 0.5 $ and  $\lambda_{SDS} = 0.3 $ initially. 

\subsubsection{View Selection}
For view selection, we initially sample camera poses uniformly distributed around the avatar in the horizontal plane. Specifically, we fix the camera parameters with an elevation of 0 degrees and a distance of 5 units from the avatar. The azimuth angles are evenly sampled at 100 intervals to ensure full 360-degree coverage. Additionally, to enhance completeness and capture top-down information, we include one overhead view with a high elevation angle. In practice, the visibility of different parts of the human body can vary significantly due to factors such as pose, occlusion, and camera placement. As a result, the ratio between visible regions (e.g., front vs. back views) is not consistent across samples or subjects. To account for this variability, we treat the visibility ratio as a hyperparameter, with a default value of 50\%, which can be tuned empirically based on the characteristics of the dataset. This allows our method to adapt to datasets with different view distributions and ensures balanced optimization of both visible and occluded regions.

\subsubsection{Progressive Training}
We design a progressive training strategy in this stage, gradually diminishing the weight of SDS loss. This strategy is employed to enhance further the effectiveness and efficiency of the visible appearance reconstruction. Based on this strategy, the $\lambda_{SDS}$ is reduced gradually as follows:
\begin{equation}
\lambda_{SDS}(t) = \lambda_{SDS,0} \cdot \frac{1}{2^{\lfloor \frac{t - t_{\text{0}}}{k} \rfloor}}
  \label{eq:prog.train}
\end{equation}
where $t$ and $t_{0}$ are the current epoch and starting epoch respectively, $k$ is the interval step of changing the weight. We set $t_{0} = 100$ and $k = 100$.

\subsection{Resolution}
The video resolution for the ZJU-Mocap (revised)\cite{liu2023zero1to3} and Monocap datasets is consistently maintained at $1024\times1024$ pixels, while MVHumanNet\cite{xiong2024mvhumannet} has a resolution of $2048\times1500$ pixels. For videos collected from the internet, the resolution ranges from 720p to 1080p. However, in Stage II, Zero123 only accepts $256\times256$ as input. Therefore, for SDS loss calculation, we crop the ground truth images based on their masks and resize them to $256\times256$.

\section{More Experiments}
\subsection{Comparison with Video-based Methods \label{resultImagebased}}
As discussed in Section~4.2, directly applying Dual-space Optimization results in degraded reconstruction quality for seen views. To overcome this limitation, we introduce View Selection and Pose Injection. Tab.~\ref{Q_table} reports the reconstruction quality for seen views, showing that our method consistently achieves performance on par with, and often surpasses, other state-of-the-art approaches. Moreover, in scenarios with significant occlusion, our method demonstrates clear advantages and see Section~\ref{sec: occlusion} for more discussion.

\begin{table*}[!t]
 
\centering
\renewcommand{\arraystretch}{1.5}
\begin{tabular}{c|ccc|ccc|ccc|ccc}
\hline
\multirow{2}{*}{} & \multicolumn{3}{c|}{ZJU-Mocap(revised)} & \multicolumn{3} {c|}{Monocap}& \multicolumn{3}{c|}{MVHumanNets}  & \multicolumn{3}{c}{In-the-wild} \\ 
                  & PSNR$\uparrow$ & SSIM$\uparrow$ & LPIPS$\downarrow$ & PSNR$\uparrow$ & SSIM$\uparrow$ & LPIPS$\downarrow$ & PSNR$\uparrow$ & SSIM$\uparrow$ & LPIPS$\downarrow$ & PSNR$\uparrow$ & SSIM$\uparrow$ & LPIPS$\downarrow$ \\ \hline
HumanNeRF         &---&---&---         &30.31&0.9642&0.0328 &30.41&0.9743&0.0321 & 30.23 & 0.9756 & 0.0314\\ 
Instant-NVR       &30.50&0.9716&0.0384 &30.73&0.9763&0.0273 &---&---&---         &---&---&---\\
SplattingAvatar   &28.93&0.9682&0.0326 &29.36&0.9703&0.0296 &28.89&0.9653&0.0306 & 28.28 & 0.9693 & 0.0286\\
ExAvatar          &30.58&0.9815&0.0269 &29.68&0.9732&0.0306 &29.45&0.9682&0.0291 & 29.46 & 0.9709 & 0.0253\\
GaussianAvatar    &29.94&0.9795&0.0210 &30.98&0.9701&0.0193 &28.58&0.9616&0.0237 & 29.96 & 0.9716 & 0.0220\\
GuessTheUnseen    &29.36&0.9535&0.0365 &---&---&---         &28.86&0.9547&0.0326 & 29.35 & 0.9685 & 0.0256 \\
Ours              &29.82&0.9786&0.0219 &30.56&0.9698&0.0216 &28.49& 0.9603&0.0259 & 29.76 & 0.9712 & 0.0222\\ \hline
\end{tabular}
\caption{Quantitative evaluation on seen views for ZJU-Mocap(revised), MVHumanNet, Monocap datasets, and In-the-wild dataset.\label{Q_table}}
\vspace{-5mm}
\end{table*}

\subsection{Comparison with Image-based Methods \label{resultImagebased}}

In this section, we compare our method with SIFU~\cite{Zhang2024SIFU}, SITH~\cite{ho2024sith}, and ELICIT~\cite{huang2022elicit}, all of which are single-image reconstruction techniques designed to synthesize unseen parts of human avatars.

\begin{figure}[!t]
  \centering
  \includegraphics[width=1\linewidth]{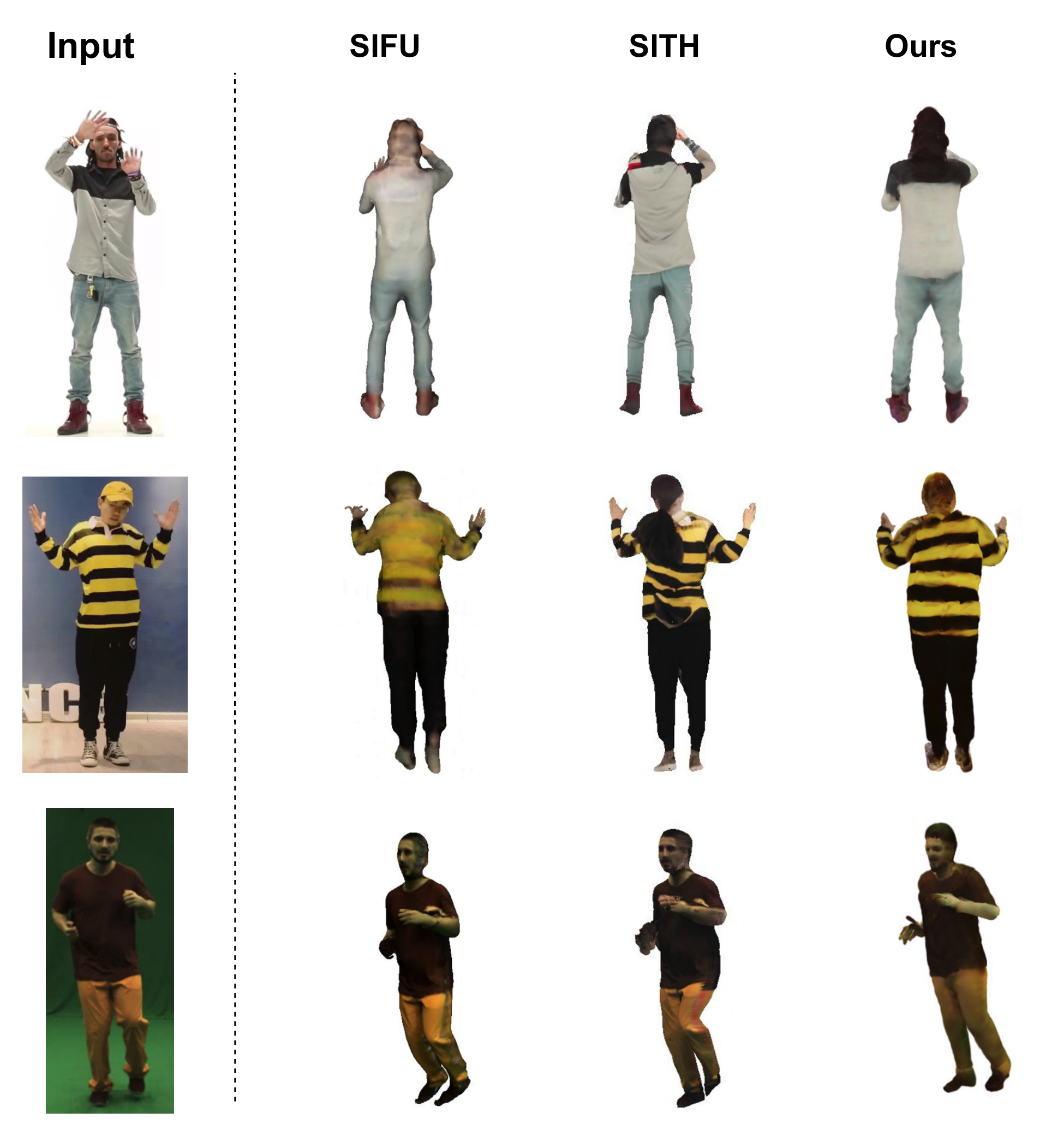}

  \caption{Qualitative comparison results with SIFU~\cite{Zhang2024SIFU} and SITH~\cite{ho2024sith}.}
    \vspace{-5mm}
  \label{fig:SIFU}
\end{figure}

SIFU proposes an approach to reconstruct clothed human avatars from single images. Qualitatively, as shown in Fig.~\ref{fig:SIFU}, this method can reconstruct decent geometry but fails to synthesize the texture of unseen parts of humans. SITH, similar to SIFU, is a method for single-image reconstruction. SITH can predict the texture of unseen parts of humans, but their generated textures contain unrealistic artifacts.

\begin{figure}[!t]
  \centering
  \includegraphics[width=1\linewidth]{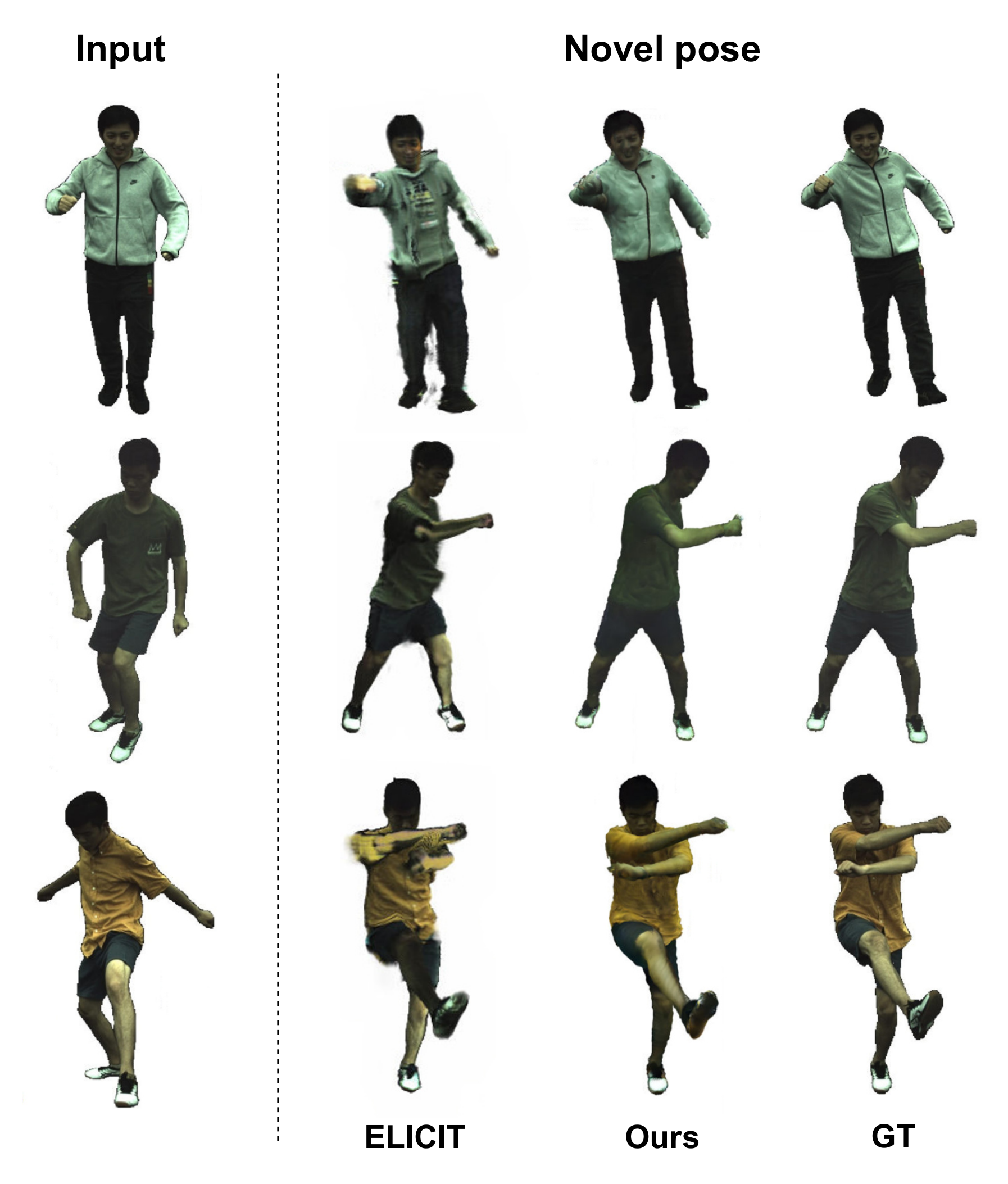}

  \caption{Qualitative comparison results with ELICIT~\cite{huang2022elicit} on novel poses.}

  \label{fig:ELICIT}
\end{figure}

ELICIT is a generative model that takes one image and a motion sequence as input to generate an animatable avatar. Qualitative results are shown in Fig.~\ref{fig:ELICIT}. For a fair comparison, since our method takes an image sequence as input, we are comparing the quality by synthesizing a novel pose that is not included in our inputs. Even though ELICIT can predict the unseen parts of humans, it shows blurred edges and floating artifacts while applying motions. Because only one image is used as input for ELICIT, the texture cannot be adapted to novel poses dynamically. In contrast, our method associates texture to different body parts across frames and can predict the correct texture for unseen parts robustly.

In Tab.~\ref{tab:exp}, we present the quantitative evaluation results. SIFU and SITH were tested on the Monocap dataset, while ELICIT was evaluated on the ZJU-Mocap(revised) dataset. The results demonstrate that our method consistently achieves superior performance compared to the state-of-the-art approaches, underscoring its efficacy and robustness.

\begin{table}[!t]
\centering
\renewcommand{\arraystretch}{1.5}

\begin{tabular}{c|c|ccc}
\hline
Dataset                    & Method & PSNR$\uparrow$&SSIM$\uparrow$&LPIPS$\downarrow$  \\ \hline
\multirow{3}{*}{MVHumanNet}   & SIFU   &19.29&0.9486&0.0706\\
                            & SITH   &19.68&0.9462&0.0699\\
                        &\textbf{Ours} & \textbf{20.98} & \textbf{0.9517} & \textbf{0.0553} \\ \hline  
\multirow{3}{*}{Monocap}   & SIFU   &18.96&0.9406&0.0659\\
                            & SITH   &19.06&0.9428&0.0673\\
                        &\textbf{Ours} & \textbf{21.16} & \textbf{0.9532} & \textbf{0.0549} \\ \hline  
\multirow{2}{*}{ZJU-Mocap(revised)} & ELICIT &19.23&0.9456&0.0689\\
                           &\textbf{Ours} & \textbf{20.82} & \textbf{0.9552} & \textbf{0.0569}  \\ \hline
\end{tabular}
\caption{Quantitative evaluation on MVHumanNet, ZJU-Mocap(revised), and Monocap datasets.}
  \label{tab:exp}
\end{table}

\subsection{Occlusion-aware Reconstruction}
\label{sec: occlusion}

Reconstructing occluded regions remains a significant challenge in full-body avatar modeling, particularly when parts of the subject are not visible in the input views. Our method addresses this issue by leveraging canonical-space priors and Dual-space Optimization (SDS) to infer and recover plausible appearance and geometry in occluded areas.

We identify three common scenarios of occlusion in video data. The first scenario involves transient occlusions, where the occluded region is visible in other frames of the video. This is the simplest case, as our method utilizes the canonical space to fuse information from well-captured frames, enabling high-fidelity reconstruction of the occluded frame without quality degradation.

\begin{figure}[tb]
\centering
\includegraphics[width=\linewidth]{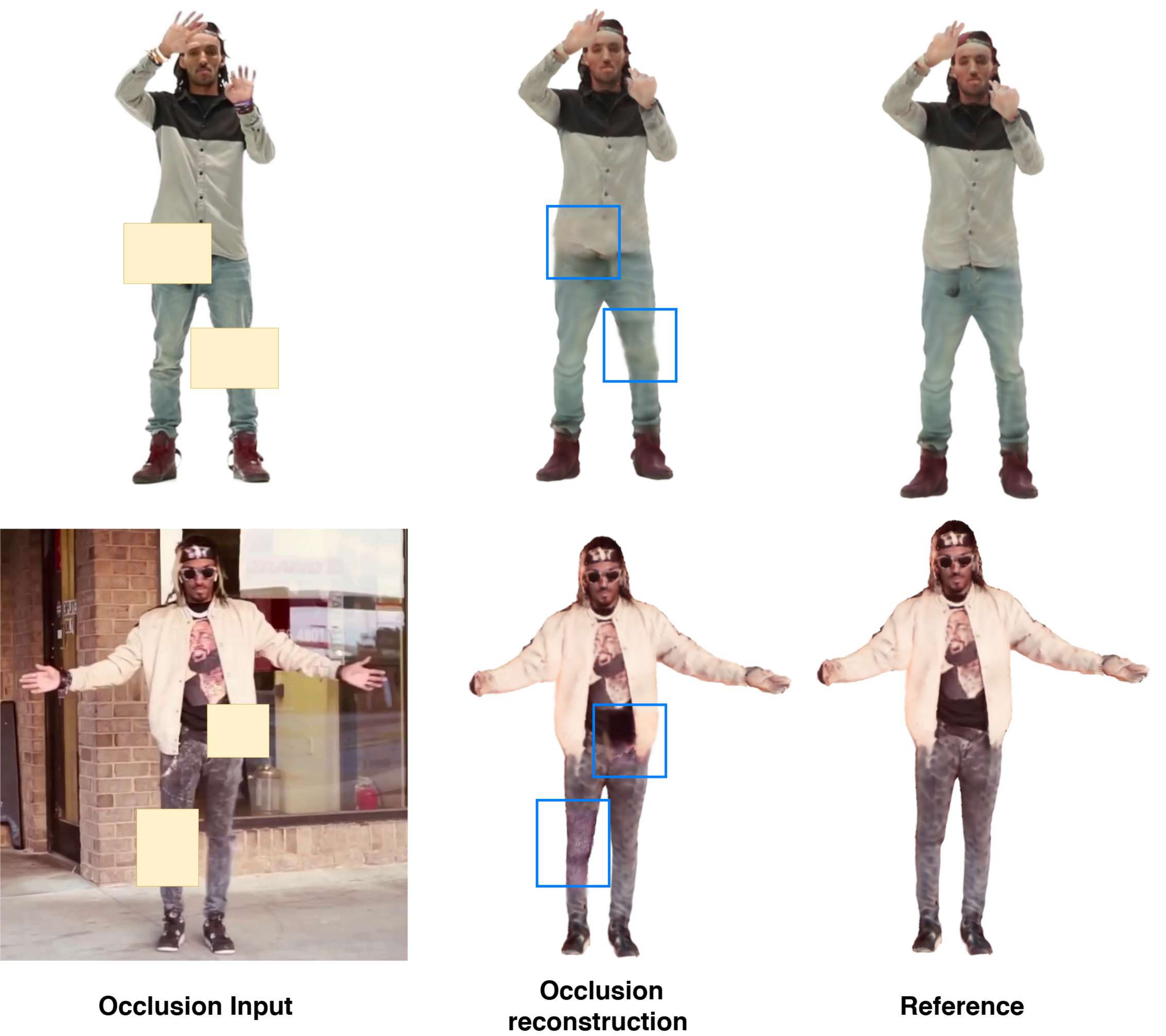} % Reduce the figure size so that it is slightly narrower than the column.

\caption{Demonstration of occlusion-aware reconstruction.}
\label{fig: occlusion}

\end{figure}

The second scenario involves partial occlusion, where a small region of the body remains blocked throughout the entire video. In such cases, our method uses a visibility map to identify the occluded regions and leverages the visible body parts as guidance. It then applies Dual-space optimization to infer the texture and geometry of the occluded areas. By combining those techniques, the model is able to reconstruct visually plausible and coherent appearances for the occluded regions, as shown in Fig.~\ref{fig: occlusion}.

The third scenario involves severe occlusion, where a large portion of the body—such as an entire leg or half of the torso—is consistently blocked throughout the video. In this case, accurate body pose fitting becomes highly challenging, as existing methods can only align to visible regions. Although one possible solution is to use Zero123 to hallucinate unseen body parts from a front-view reference image, it requires that the missing region be present in the reference. If the occluded parts are absent in both the input and reference views, our method cannot recover them. Therefore, our current pipeline is not capable of handling such extreme occlusions.

\label{occlusion}

\subsection{Efficiency} Our method, with its two-stage training process, requires approximately 3 training hours, significantly outperforming HumanNeRF, which demands about 10 hours of training on the same device. Furthermore, our method can achieve almost real-time rendering speed at 18 FPS. But Instant-NVR and HumanNeRF can only render with 2 FPS and 7 FPS respectively. In comparison to 3DGS-based approaches, our method is slower than GuessTheUnseen~\cite{lee2024gtu} (150 FPS) but operates at a similar speed to GaussianAvatar~\cite{hu2023gaussianavatar}.

The training process takes 300 epochs, corresponding to approximately 60,000 iterations, which is slower compared to ExAvatar~\cite{moon2024exavatar}(10,000 iterations), SplattingAvatar~\cite{shao2024splattingavatar}(15,000 iterations), and GaussianAvatar~\cite{hu2023gaussianavatar}(30,000 iterations). This increased training time is primarily due to the slower convergence of the SDS loss and the additional supervision from the normal map loss.

\subsection{Novel Poses Animation}
Our method aligns the generated Gaussian human avatars with the SMPL model, enabling us to animate the reconstructed avatar with novel poses, as shown in Fig.~\ref{fig:anim}. Please refer to the accompanying video for dynamic results.
\begin{figure}[tb]
  \centering
  \includegraphics[width=\linewidth]{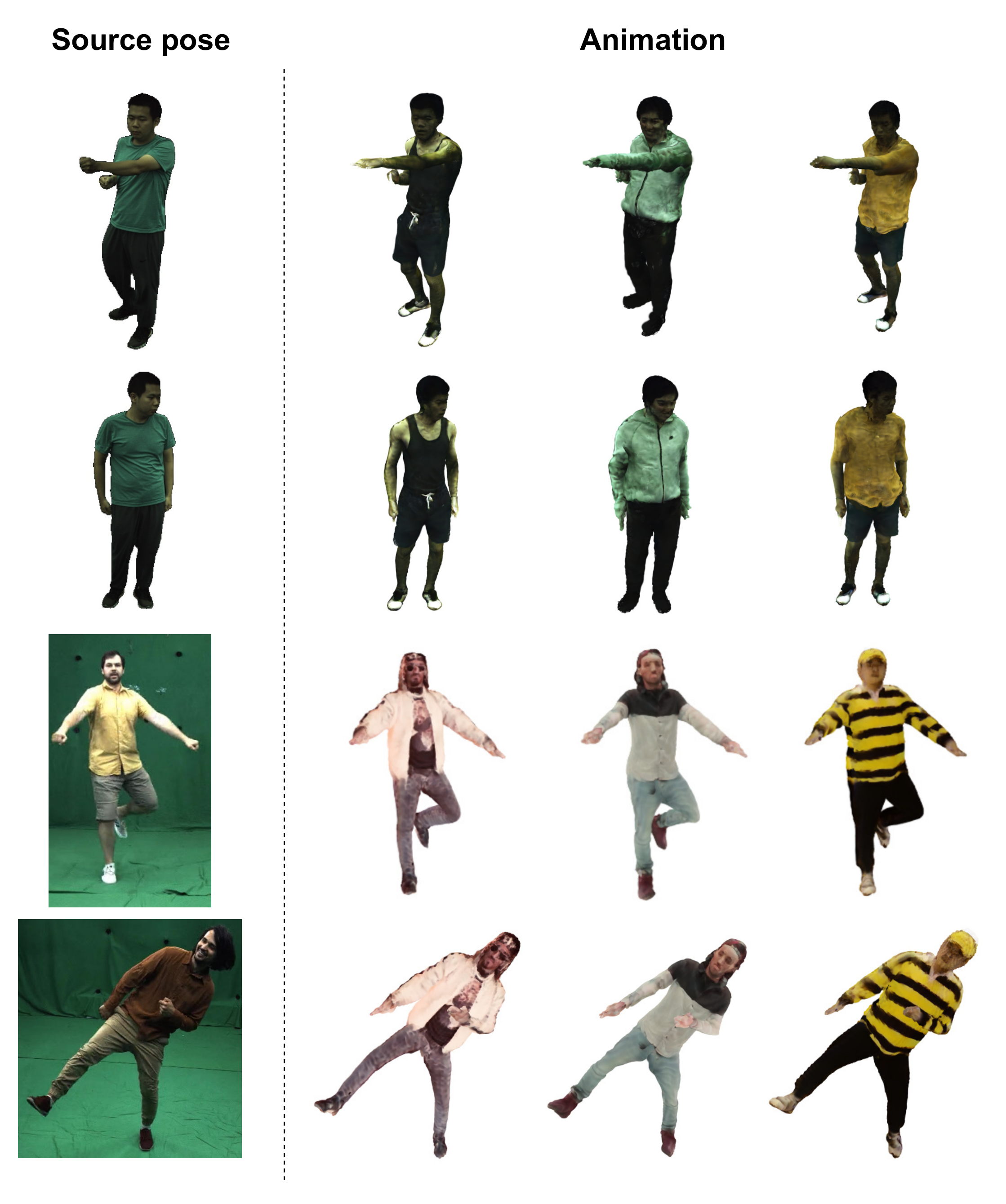}

  \caption{Avatar animations with novel poses.}
    
  \label{fig:anim}
\end{figure}

\begin{figure}[!t]
\centering
\includegraphics[width=\linewidth]{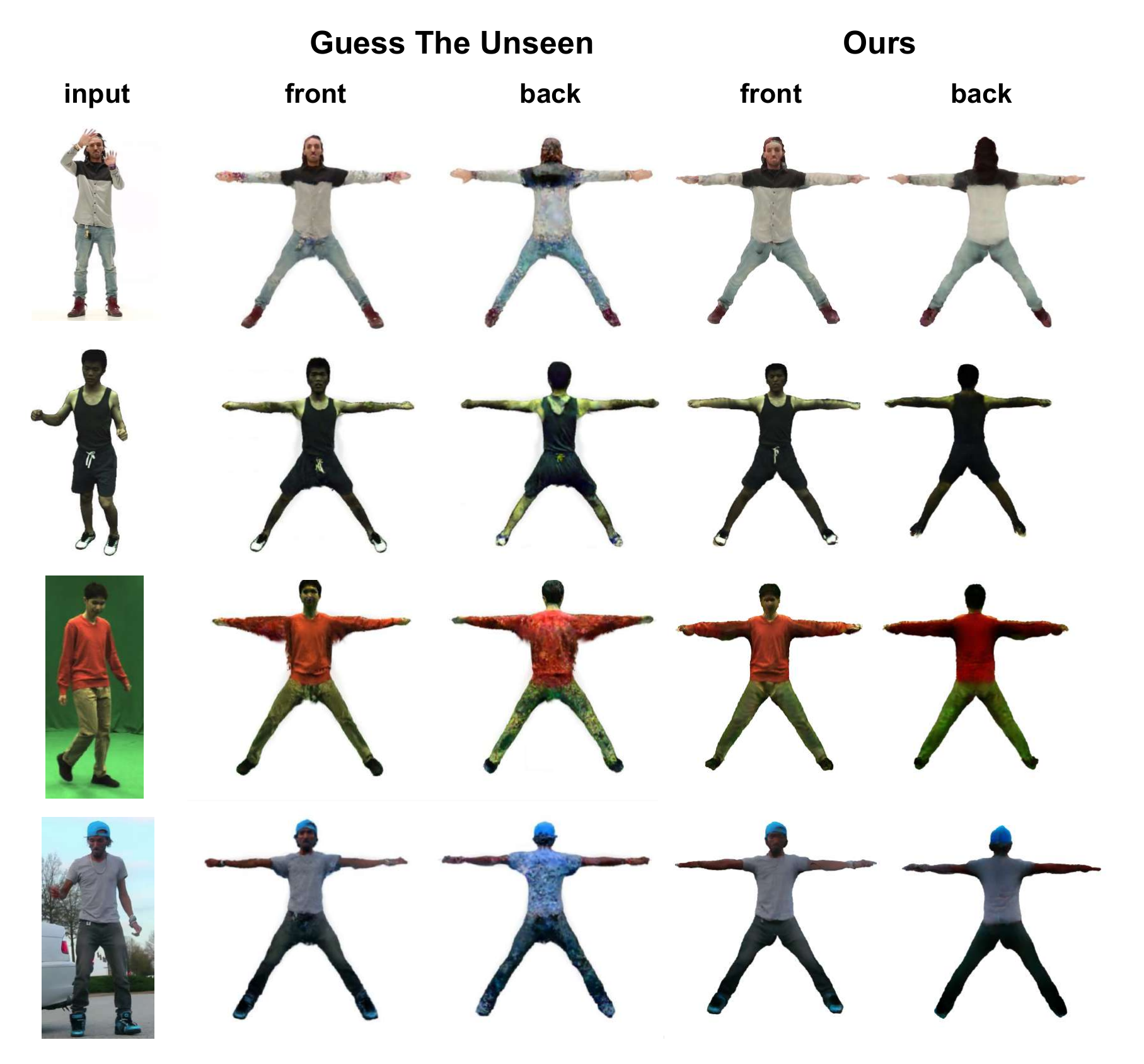} 

\caption{More qualitative comparison with GuessTheUnseen~\cite{lee2024gtu}.}
\label{fig: qualitati.ve2}

\end{figure}

\begin{figure*}[!t]
  \centering
  \includegraphics[width=\textwidth]{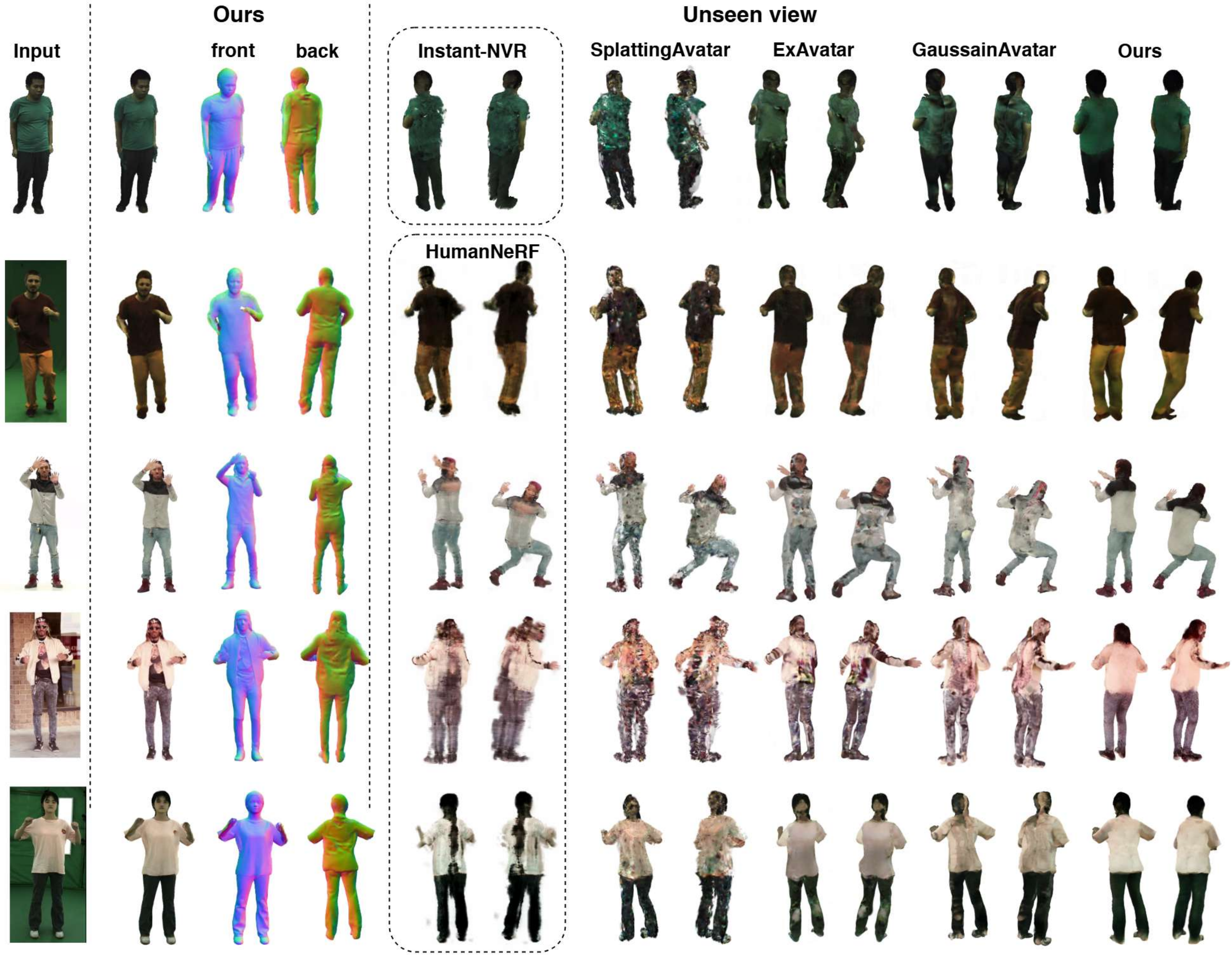}

  \caption{More qualitative comparison with HumanNeRF~\cite{weng2022humannerf}, Instant-NVR~\cite{instant_nvr},  SplattingAvatar~\cite{shao2024splattingavatar}, ExAvatar~\cite{moon2024exavatar} and GaussianAvatar~\cite{hu2023gaussianavatar}.}
    
  \label{fig:anim}
\end{figure*}

% \ifCLASSOPTIONcompsoc

% %  \section*{Acknowledgments}
% \else

%   \section*{Acknowledgment}
% \fi
% % The authors would like to thank the anonymous reviewers for their constructive comments and suggestions. Zhang, Li, and Guo were partially supported by a grant from National Science Foundation (2007661) and research gifts from Samsung Research America. Zeng was partially supported by NSFC (No.62072382), and Fundamental Research Funds for the Central Universities, China (No.20720190003). The opinions expressed are solely those of the authors, and do not necessarily represent those of the funding agencies.

% \ifCLASSOPTIONcaptionsoff
%   \newpage
% \fi

% \balance
\bibliographystyle{IEEEtran}
\bibliography{TVCG}
\vspace{-16mm}
\begin{IEEEbiography}[{\includegraphics[width=1in,height=1.25in,trim=70 70 80 70, clip,keepaspectratio]{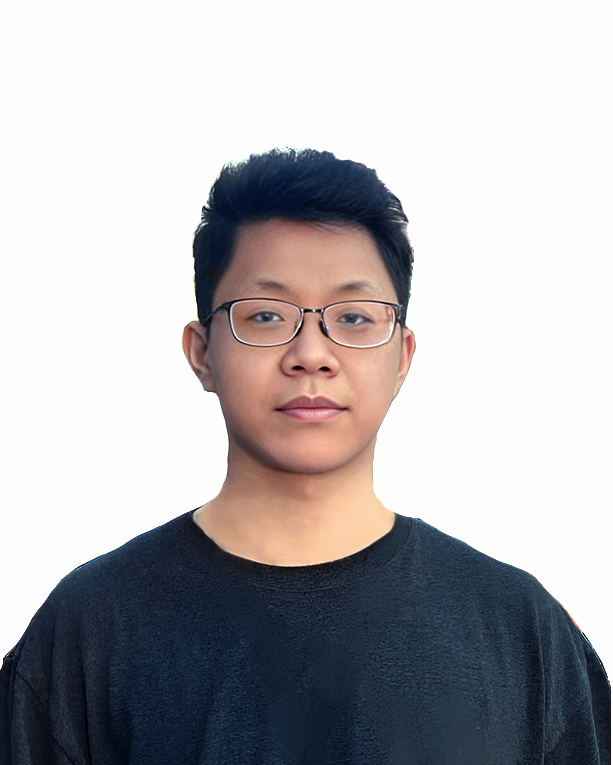}}]{Zilong~Wang} is a Ph.D. candidate at the University of Texas at Dallas, supervised by Prof. Xiaohu Guo. He received his B.S. degree in software engineering in 2020 from Northwest University(China) and M.S. degree in software engineering in 2022 from the University of Texas at Dallas. 
His research interests include human reconstruction and animation, computer graphics, computer vision, and deep learning.
\end{IEEEbiography}
\vspace{-16mm}
\begin{IEEEbiography}[{\includegraphics[width=1in,height=1.25in,trim=70 40 80 70,clip,keepaspectratio]{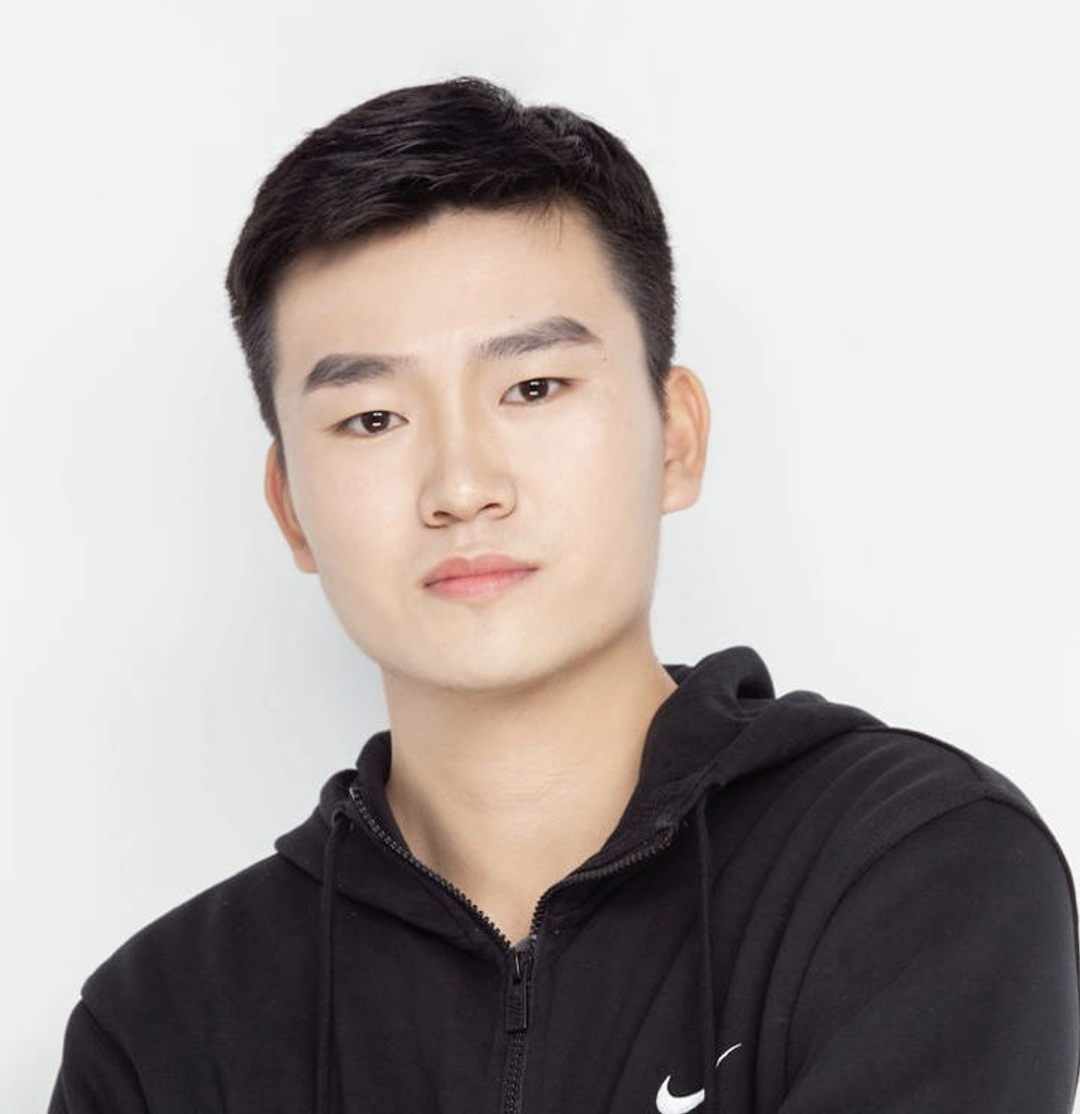}}]{Zhiyang (Frank) Dou} is a Ph.D. student at MIT CSAIL. He previously completed his M.Phil. in the Computer Graphics Group at The University of Hong Kong, where he was advised by Prof. Wenping Wang and Prof. Taku Komura. %Zhiyang earned his B.Eng. degree with honors from Shandong University, where he was advised by Prof. Shiqing Xin.
Zhiyang’s research focuses on shape recovery and generation, character animation, geometric modeling, and the analysis of human behavior, emphasizing the intersection of artificial intelligence, computer graphics and computer vision.
\end{IEEEbiography}
\vspace{-15mm}
\begin{IEEEbiography}[{\includegraphics[width=1in,height=1.25in,trim=40 40 40 40,clip,keepaspectratio]{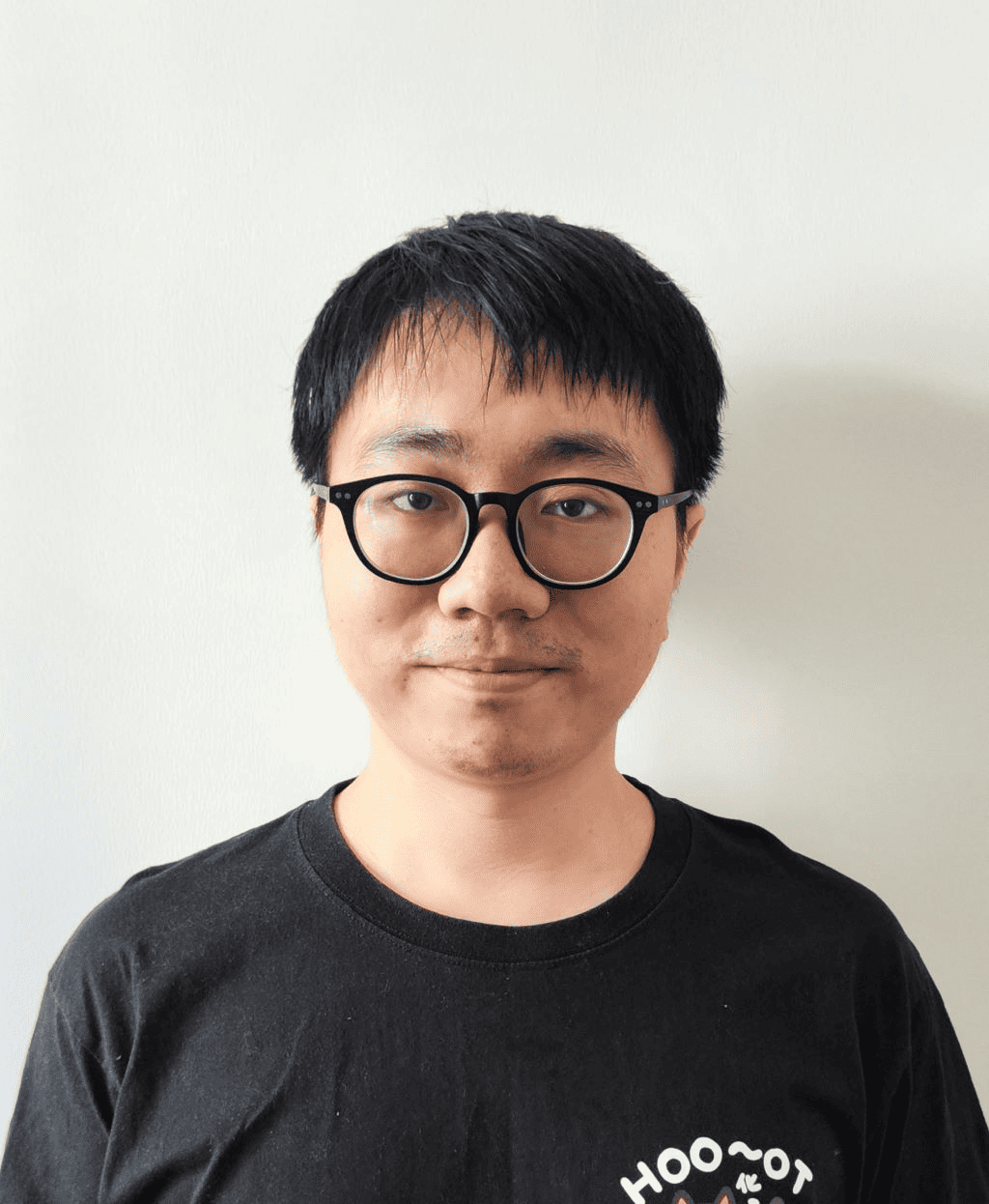}}]{Yuan~Liu} is an assistant professor at HKUST. He received his PhD degree in the University of Hong Kong in 2024. His research mainly concentrates on 3D vision and graphics. I currently work on topics about 3D AIGC including neural rendering, neural representations, and 3D generative models.
\end{IEEEbiography}
\vspace{-15mm}
\begin{IEEEbiography}[{\includegraphics[width=1in,height=1.25in,clip,keepaspectratio]{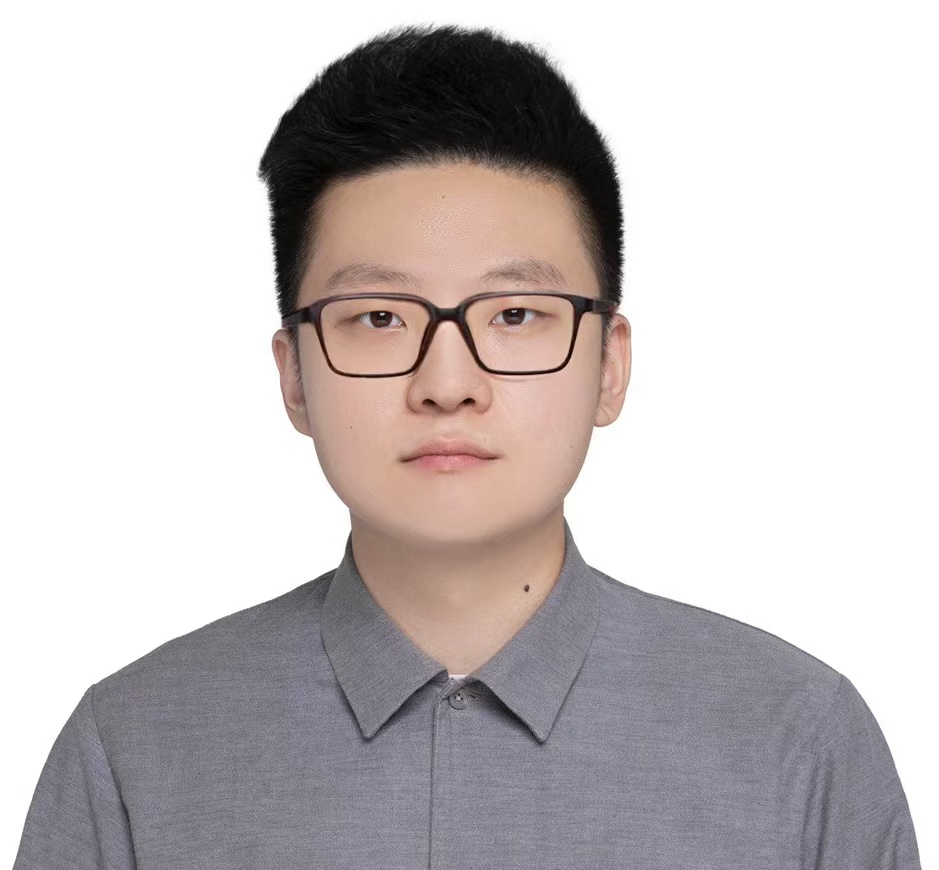}}]{Cheng~Lin} received his Ph.D. from The University of Hong Kong (HKU), advised by Prof. Wenping Wang. He visited the Visual Computing Group at Technical University of Munich (TUM), advised by Prof. Matthias Nießner. Before that, he completed his B.E. degree at Shandong University. His research interests include geometric modeling, 3D vision, shape analysis, and computer graphics.
\end{IEEEbiography}
\vspace{-16mm}
\begin{IEEEbiography}[{\includegraphics[width=1in,height=1.25in,clip,keepaspectratio]{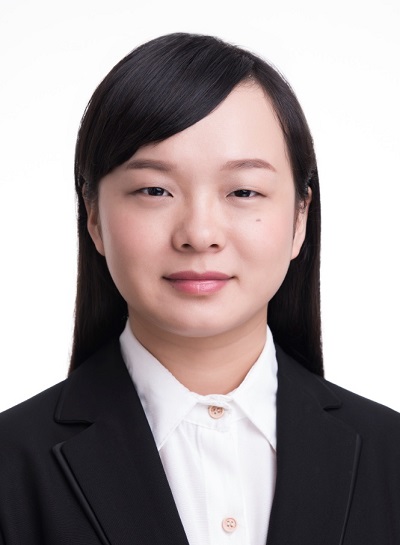}}]{Xiao~Dong} is an assistant Professor in the Department of Computer Science, Beijing Normal-Hong Kong Baptist University. She received the BS and PhD degrees in computer science and technology from Xiamen University, in 2013 and 2022, respectively. Her research interests include computer graphics, computer vision and deep learning.
\end{IEEEbiography}
\vspace{-16mm}
\begin{IEEEbiography}[{\includegraphics[width=1in,height=1.25in,clip,keepaspectratio]{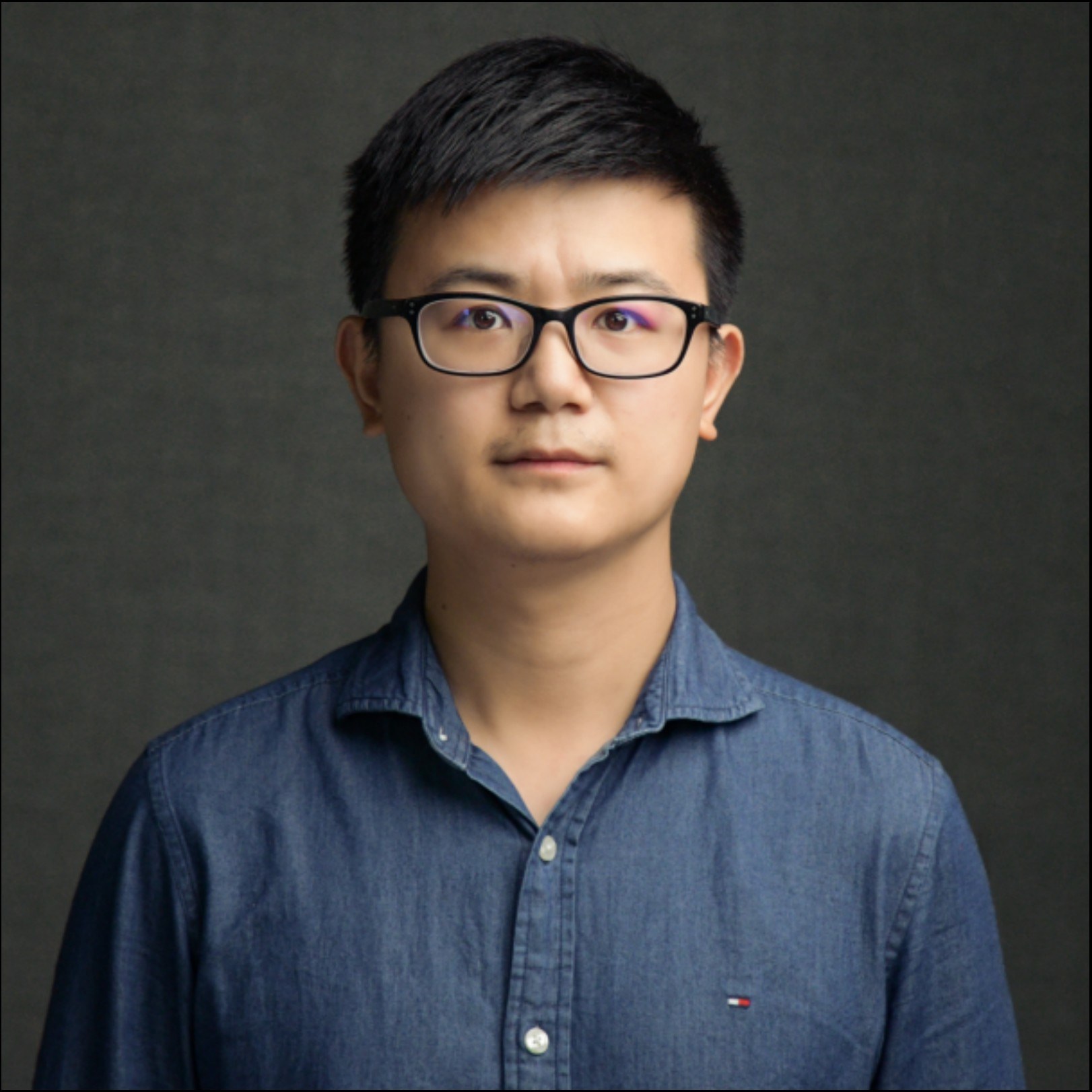}}]{Yunhui~Guo} is an assistant professor in the Department of Computer Science at the University of Texas at Dallas. Previously, he was a postdoctoral researcher at UC Berkeley/ICSI. He earned his PhD in Computer Science from the University of California, San Diego. His research interests include machine learning and computer vision, with a focus on developing intelligent agents that can continuously learn, dynamically adapt to evolving environments without forgetting previously acquired knowledge, and repurpose existing knowledge to handle novel scenarios.
\end{IEEEbiography}
\vspace{-16mm}
\begin{IEEEbiography}[{\includegraphics[width=1in,height=1.25in,clip,keepaspectratio]{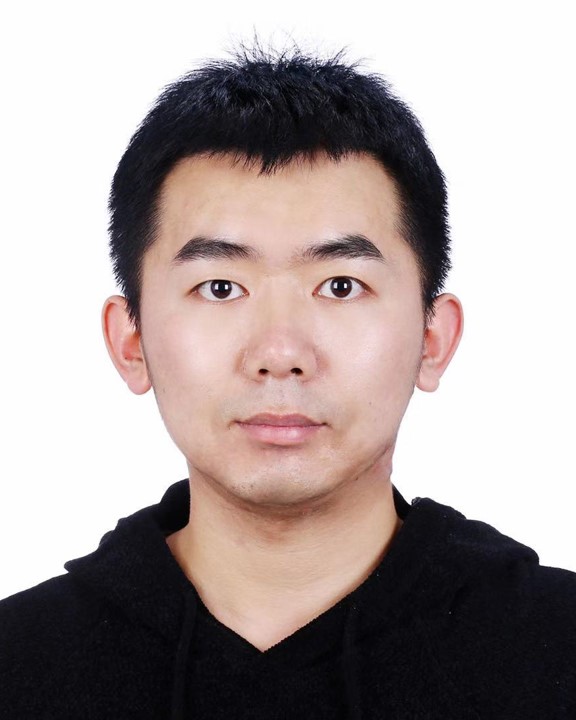}}]{Chenxu~Zhang} is a Research Scientist at the Intelligent Creation Lab, ByteDance. He completed his Ph.D. degree in Computer Science from the University of Texas at Dallas in 2023. He received his B.S. degree in Software Engineering in 2015 and M.S. degree in Computer Science in 2018, both from Beihang University. His research interests include computer graphics, computer vision, and deep learning.
\end{IEEEbiography}
\vspace{-16mm}
\begin{IEEEbiography}[{\includegraphics[width=1in,height=1.25in,clip,keepaspectratio]{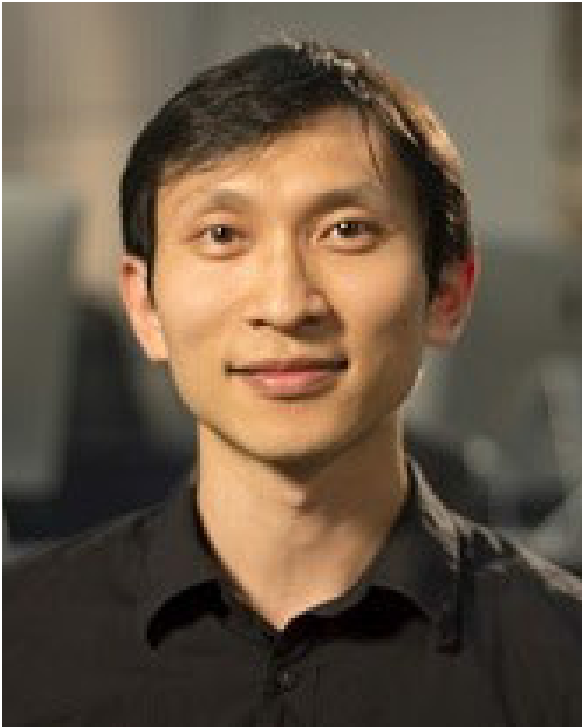}}]{Xin~Li} (Senior Member, IEEE) received the B.E. degree in computer science from the University of Science and Technology of China in 2003 and the M.S. and Ph.D. degrees in computer science from the State University of New York at Stony Brook in 2005 and 2008, respectively. He is currently a Professor with the Section of Visual Computing and Creative Media, School of Performance, Visualization, and Fine Arts, Texas A\&M University. His
research interests include geometric and visual data computing, processing, and understanding, computer vision, and virtual reality.
\end{IEEEbiography}
\vspace{-16mm}
\begin{IEEEbiography}[{\includegraphics[width=1in,height=1.25in,clip,keepaspectratio]{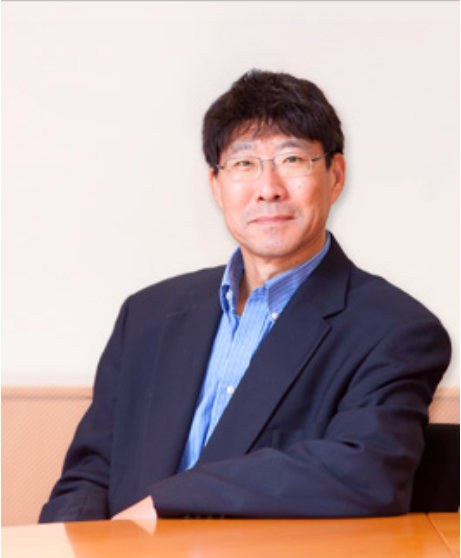}}]{Wenping~Wang} (Fellow, IEEE) received the
Ph.D. degree in computer science from the University of Alberta. He is a Professor of computer science at Texas A\&M University. His research interests include computer graphics, computer vision, robotics, medical image processing, and geometric computing. He has been a journal associate editor of ACM Transactions on Graphics, IEEE Transactions on Visualization and Computer Graphics, Computer Aided Geometric Design, and Computer Graphics Forum (CGF). He has chaired a number of international conferences, including Pacific Graphics, ACM Symposium on Physical and Solid Modeling (SPM), SIGGRAPH and SIGGRAPH Asia. Prof. Wang received the John Gregory Memorial Award for his contributions to geometric modeling.
\end{IEEEbiography}
\vspace{-15mm}
\begin{IEEEbiography}[{\includegraphics[width=1in,height=1.25in,clip,keepaspectratio]{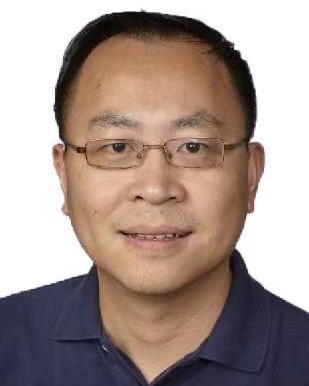}}]{Xiaohu~Guo} is a Full Professor of Computer Science at the University of Texas at Dallas. He received his Ph.D degree in Computer Science from Stony Brook University, and a B.S degree in Computer Science from the University of Science and Technology of China. His research interests include computer graphics, computer vision, medical imaging, with an emphasis on geometric modeling and processing, as well as body and face modeling problems. He received the prestigious NSF CAREER Award in 2012 and SIGGRAPH 2023 Best Paper Award. He has been serving on the journal editorial boards of IEEE TVCG, TMM, TCSVT, GMOD, CAVW, and on the executive committee of Solid Modeling Association.
\end{IEEEbiography}

\end{document}